\documentclass{article}

\usepackage{array}
\usepackage{PRIMEarxiv}
\usepackage{natbib}
\usepackage[utf8]{inputenc} 
\usepackage[T1]{fontenc}    
\usepackage{hyperref}       
\usepackage{url}            
\usepackage{booktabs}       
\usepackage{amsfonts}       
\usepackage{nicefrac}       
\usepackage{microtype}      
\usepackage{lipsum}
\usepackage{graphicx}
\graphicspath{{media/}}     
\usepackage{paralist}
\usepackage{algorithm}
\usepackage{algorithmic}
\usepackage{multirow}
\usepackage{overpic}
\usepackage{ulem}
\usepackage{comment}
\usepackage{xcolor}
\newcommand{{\ourmethod}}[0]{HAAD}
\newcommand{{\Supp}}[0]{Supp. Material}

\newcommand{\encoder}{\mathcal{E}}
\newcommand{\decoder}{\mathcal{D}}

\newcommand{\expec}{\mathbb{E}}
\newcommand{\model}{\epsilon_\theta}

\newcommand{\conditioner}{\tau_\theta}

\usepackage{amsmath}

\title{An \textit{h}-space Based Adversarial Attack for Protection Against Few-shot Personalization}

\author{
  Xide Xu \\
  Computer Vision Center \\
  Universitat Autònoma de Barcelona \\
  Barcelona, Spain\\
  \texttt{xide@cvc.uab.cat} \\
   \And
  Sandesh Kamath \\
  Computer Vision Center \\
  Universitat Autònoma de Barcelona \\
  Barcelona, Spain\\
  \texttt{skamath@cvc.uab.cat} \\
   \And
  Muhammad Atif Butt \\
  Computer Vision Center \\
  Universitat Autònoma de Barcelona \\
  Barcelona, Spain\\
  \texttt{mabutt@cvc.uab.cat} \\
    \And
  Bogdan Raducanu \\
  Universitat Autònoma de Barcelona \\
  Barcelona, Spain\\
  \texttt{bogdan@cvc.uab.cat} \\
}

\begin{document}
\maketitle

\begin{abstract}
The versatility of diffusion models in generating customized images from few samples raises significant privacy concerns, particularly regarding unauthorized modifications of private content. This concerning issue has renewed the efforts in developing protection mechanisms based on adversarial attacks, which generate effective perturbations to poison diffusion models. Our work is motivated by the observation that these models exhibit a high degree of abstraction within their semantic latent space (termed `\textit{h}-space'), which encodes critical high-level features for generating coherent and meaningful content. In this paper, we propose a novel anti-customization approach, called {\ourmethod} (\textbf{\textit{h}}-space based \textbf{A}dversarial \textbf{A}ttack for \textbf{D}iffusion models) that leverages adversarial attacks to craft perturbations based on the \textit{h}-space that can efficiently degrade the image generation process. Building upon {\ourmethod}, we further introduce a more efficient variant, {\ourmethod}-KV, that constructs perturbations solely based on the KV parameters of the \textit{h}-space. This strategy offers a stronger protection, that is computationally less expensive. Despite their simplicity, our methods outperform state-of-the-art adversarial attacks, highlighting their effectiveness.

\textcolor{red}{[Warning: This paper may contain images that could produce visual discomfort.]}
\end{abstract}


\section{Introduction}
\label{introduction}

Diffusion Models \cite{ho2020ddpm,song2021ddim,song2021score} are the current state of the art for image generation, outperforming GANs in terms of image quality and mode coverage \cite{dhariwal2021neurips,song2021maxlikelihood}. With the introduction of large-scale diffusion models, \cite{rombach2022ldm,saharia2022imagen,ramesh2022dalle2} able to generate intricate details and complex patterns via textual prompts, we achieve unmatched accuracy and robustness towards generating high-fidelity images. Image editing \cite{hertz2023p2p,mokady2023nulltext}, image-to-image translation \cite{saharia2022palette}, text-to-3D images synthesis \cite{poole2023dreamfusion,xu2023dream3d,tang2023makeit3D}, video generation \cite{ho2022vdm,blattmann2023svd,gupta2023video}, anomaly detection in medical images \cite{wolleb2022miccai}, etc. are few of the many applications of diffusion models. One application that has become extremely popular due to its versatility and ease of use is the customization of diffusion models with personal content \cite{gal2022ti,ruiz2023db,kumari2023cd}. These models have the ability to create personalized content from few user images by fine-tuning a pre-trained diffusion model (e.g. Stable Diffusion) in order to learn how to bind an unique token to a novel concept. Consequently, we can generate novel views in different contexts and visualize them under different artistic styles.

Few-shot image customization leverages diffusion models to create user-tailored content, adapting the output to fit particular preferences. This opens new ways of creating a more engaging and meaningful interaction with AI systems. The ability of diffusion models to learn and adapt to specific user inputs has revolutionized the field of personalized content creation, making it more accessible and impactful across various industries, such as visual arts (artworks), virtual reality, gaming, and e-commerce. While these customization approaches are powerful tools for generating user-specific content, they also represent significant privacy risks. Malicious individuals could exploit the vulnerability of this technology to produce and spread deceptive images (`deep fakes') that are visually indistinguishable from genuine ones \cite{masood2023deepfake}. The negative effects of privacy risks induced by deep fakes could span from information leakage, unauthorized reproduction of artwork \cite{shan2023glaze,shan2024nightshade} to extreme cases where it could severely impact an individual's personal life and reputation \cite{chesney2019deepfake}.

\begin{figure}[ht]

\begin{center}

\centerline{\includegraphics[width=0.85\columnwidth]{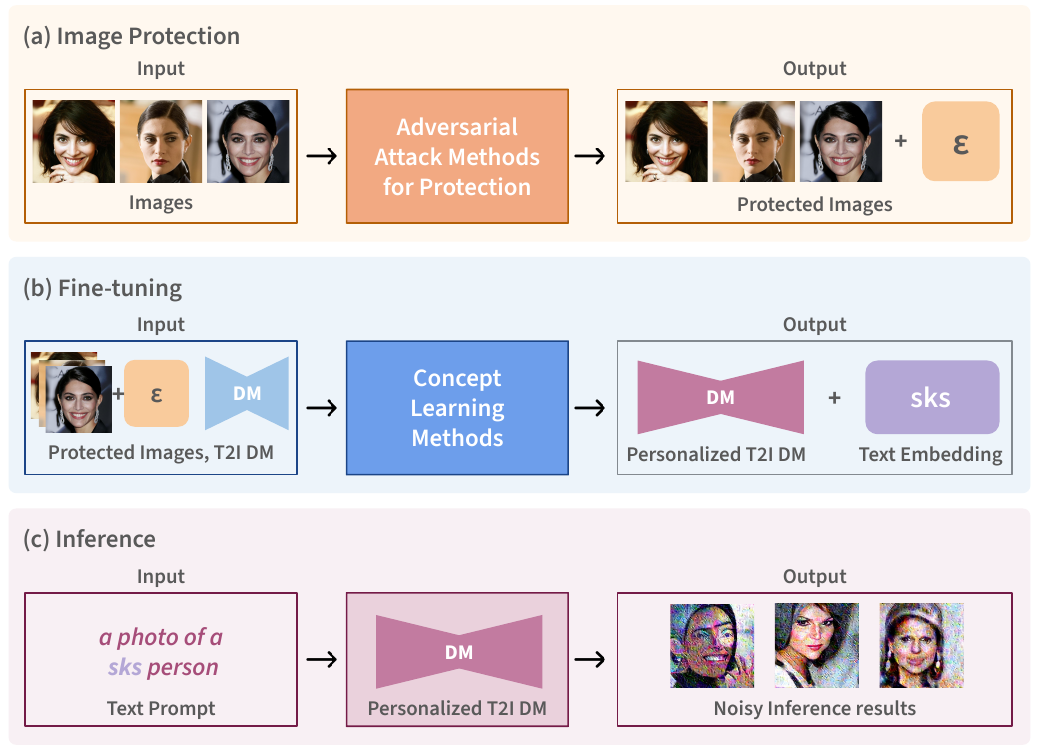}}

\caption{A step-by-step explanation of protection against customization by few-shot personalization methods using adversarial perturbations: a) Starting with a set of clean images, these methods obtain the protected images using adversarial attack methods (e.g. PGD); b) During fine-tuning, a diffusion model is personalized with these protected images leading to a poisoned model; c) at inference, the poisoned model will not be able to generate good customized images.}
\label{fig:defense-scenario}
\end{center}
\vskip -0.2in
\end{figure}

In order to prevent these privacy risks, there are currently some efforts that explore protection mechanisms that can prevent against the malicious use of customized diffusion models \cite{zhang2025surveyadvattack}. The main idea of the anti-customization methods is to obtain a protected image i.e an adversarial example, from a text-to-image (T2I) diffusion model, with the well-known Projected Gradient Descent (PGD) algorithm \cite{madry2018pgd}. This protected image when used by an attacker to train a diffusion model, leads to its poisoning, thereby successfully preventing it from generating good images. For a pictorial representation of this flow, refer to Figure \ref{fig:defense-scenario}. Recently, both targeted and untargeted approaches have been proposed for attacking the T2I diffusion models with adversarial examples \cite{salman2023icml,liang2023advdm}. Targeted attacks are designed to disrupt a model's functionality by forcing it to produce a specific, predetermined output (e.g. by altering a generated image to match a specific pattern) \cite{zheng2025targeted}. On the other hand, untargeted attacks \cite{vanle2023antidb,xu2024caat,xue2024iclr} disrupt the overall model's functionality, without guiding it towards a specific output. One of the limitations of targeted attacks is that they may leave visible traces of the specific pattern in the protected image, which a skillful person could potentially analyze in order to purify and recover the original output. 

This remarkable capability of diffusion models to generate user-specific content is rooted in their underlying semantic
latent space \cite{kwon2023latentspace}, termed `\textit{h}-space', which is responsible for generating high-quality images, coherent with the user's input. \textit{h}-space represents the deep features from the middle-block of the U-Net architecture within the denoiser component of the diffusion model (see Fig. \ref{fig:haad-block-diagram}). Leveraging this high degree of abstraction within the \textit{h}-space, we propose an adversarial attack which serves as an anti-customization method, named {\ourmethod} (\textbf{\textit{h}}-space based \textbf{A}dversarial \textbf{A}ttack for \textbf{D}iffusion models). The imperceptible perturbations introduced by our approach disrupts the model's ability to maintain consistency with user-specific input, effectively interfering with the generation of personalized content and hence, providing a strong protection to the personal data. Building on prior works \cite{kumari2023cd,chefer2023ae} that emphasize the critical role of cross-attention mechanisms in guiding semantic alignment and content fidelity within diffusion models, we further construct a more efficient variant of our method, termed {\ourmethod}-KV. Instead of considering the entire \textit{h}-space, {\ourmethod}-KV focuses solely on disrupting the key (\textit{K}) and value (\textit{V}) parameters within the cross-attention layer of the U-Net's \textit{h}-space. This targeted intervention leverages the high semantic influence of the attention pathway while requiring significantly fewer parameter($\sim$0.05\%) updates w.r.t. HAAD. As a result, HAAD-KV achieves enhanced attack performance with reduced computational overhead.

We summarize our contributions in the list below:

\begin{itemize}

    \item we propose HAAD, a simple yet effective and robust approach for privacy protection by constructing a perturbation based on the `\textit{h}-space' of diffusion models. 
    \item additionally, we introduce a more efficient variant, by focusing only on the KV parameters of the cross-attention layer ({\ourmethod}-KV), which provides increased protection at a small fraction of computational cost.
    \item we show through extensive comparison with several models and validation datasets that our approach and its variant not only outperform state-of-the-art methods based on adversarial attack, but also present increased robustness against a variety of purification strategies.

\end{itemize}

\section{Related Work}

\subsection{Personalized Diffusion Models}

The personalization of text-to-image (T2I) diffusion models has gained significant attention due to their ability to generate user-specific content. This process typically requires only a few reference images (commonly 3–5, or around 20 for high-quality face images) and trains a model to associate a unique identifier with the target concept, bridging the gap between generic AI-generated imagery and highly customized visual content. Early approaches to personalization focused on text embedding adaptation rather than model fine-tuning. Textual Inversion \cite{gal2022ti} pioneered this direction by learning a compact text embedding vector to represent a new concept through inverting visual examples into the pretrained model's text space. While parameter-efficient, this method relies heavily on the pretrained knowledge and struggles to capture complex visual details compared to fine-tuning. By better fine-tuning the model, DreamBooth \cite{ruiz2023db} achieves improved personalized generation. While effective, this approach is computationally expensive and risks overfitting. To mitigate this, Custom Diffusion (CD) \cite{kumari2023cd} introduces and fine-tunes additional cross-attention layers while keeping the original model frozen. This selective adaptation significantly reduces computational costs while preserving the model’s generalization, offering a more efficient and scalable solution for personalization. Another method which provides an efficient way for parameter fine-tuning is SVDiff \cite{han2023svdiff}. SVDiff uses Singular Value Decomposition (SVD) to create a low-rank approximation of weight updates potentially offering even greater parameter efficiency. More recent approaches include \cite{chen2024disenbooth,shi2024instantbooth,ruiz2024hyperdb}.

\vspace{-3mm}

\subsection{Anti-Customization of Diffusion Models}
A number of recent works have explored adversarial attacks as a protection method to counter unauthorized customization using diffusion models. Among the earliest, AdvDM \cite{liang2023advdm} introduces the idea of generating adversarial noise during training by maximizing the model’s original loss function, thereby preventing effective learning from perturbed inputs. While effective in scenarios such as artwork protection, its optimization focuses on the training loss rather than internal semantics, making the protection susceptible to adaptation by personalized models.

Other approaches, such as PhotoGuard \cite{salman2023icml}, Mist \cite{liang2023mist}, and ACE \cite{zheng2025targeted}, generate visually imperceptible noise using fixed global patterns—ranging from random noise to high-frequency Moiré textures—to guide model outputs toward predefined targets or degradation. However, these patterns operate at the pixel level, lacking alignment with the model’s internal representations. As a result, they often fail to disrupt semantics consistently in the generated content. Worse yet, they may leave visually detectable patterns. 

Several methods attempt to improve the generality and robustness of the protection by introducing perturbations at the training stage. Anti-DreamBooth \cite{vanle2023antidb} and its extension MetaCloak \cite{liu2024metacloak} inject perturbations using surrogate models, sometimes combining loss components to induce instability in the model’s optimization. However, these approaches still do not leverage the semantic abstraction capacity of diffusion models, instead relying on large-scale or iterative optimization that is costly and often model-specific.

CAAT \cite{xu2024caat} takes a more parameter-efficient route by targeting only the cross-attention layers in the customized diffusion (CD) model, noting that these components undergo the most change during personalization. While this improves efficiency, it treats the attention modules as a monolithic block and does not distinguish between different attention projections.

In this work, we overcome both semantic and efficiency limitations through two key innovations. First, we propose HAAD, which injects perturbations into the \textit{h}-space, which leads to stronger misalignment between input identity and output semantics, even under small noise budgets. Second, we introduce a more efficient variant, {\ourmethod}-KV, that perturbs only the KV parameters of the cross-attention layer within the \textit{h}-space. 
Together, these innovations result in a stronger, more resilient protection that outperforms prior state-of-the-art approaches in both effectiveness and computational efficiency.

\section{Method}

\subsection{Latent Diffusion Models}
A diffusion model consists of a forward process where the input image is disrupted by adding noise in multiple steps and a reverse (i.e. generative) process where the final image is obtained by applying multiple denoising steps. A latent diffusion model (LDM)\cite{rombach2022ldm} is a diffusion model where the diffusion processes are applied to the latent space instead of the image space. LDM consists of two main components: (i) an autoencoder ($\encoder$) that transforms an image $x$ into a latent code $z_0=\encoder(x)$, while the decoder ($\decoder$) reconstructs the latent code back to the original image such that $\decoder(\encoder(x)) \approx x$; and (ii) a diffusion model (parameterized by $\theta$), which applies the diffusion processes on the latent space,
commonly a U-Net~\cite{ronneberger2015unet} based model which can be conditioned using class labels, segmentation masks, or textual input. Let $\tau_\theta(y)$ represent the conditioning mechanism (e.g. prompt) that converts a condition $y$ into a conditioning vector and $t \in T$ be the number of steps of the diffusion process. The reconstruction loss used to train the model is : %

\begin{equation}
    \mathcal{L}_{\text{LDM}} = \expec_{z_0, y, \epsilon \sim \mathcal{N}(0, 1)} \Vert \epsilon - \model(z_{t},t, \conditioner(y)) \Vert_{2}^{2},
    \label{eq:loss}
\end{equation}

where, $\model$ is the conditional U-Net~\cite{ronneberger2015unet} that predicts the noise added in the denoising step.

\subsection{Adversarial Attacks on Diffusion Models}
Adversarial attacks compute a subtle, human-imperceptible perturbation, added to the input data which gets grossly misclassified. It exposes the brittleness of deep learning classification models which obtain super-human performance. An adversarial attack is formulated as follows: given an input $x$, obtain a perturbed input $x'$ such that :

 \begin{equation}
 \underset{x'}{argmax}\ \mathcal{L}_{\phi}(x')\; \quad
 s.t. \ ||x'-x||_{\infty} \leq \eta
 \end{equation}
\vspace{-2mm}

where, $\mathcal{L}_{\phi}$ is the classification loss function i.e. cross-entropy used to train the model with parameters $\phi$. We use the $\ell_{\infty}$-norm to control the noise budget given by $\eta$. Commonly, the strong iterative PGD algorithm \cite{madry2018pgd} is used to construct the adversarial attack. For diffusion models, the main focus of this work, the above objective remains the same: obtaining an adversarial image $x'$ from a clean image $x$ perturbed with an imperceptible noise (within $\eta$-budget). 

\begin{figure}[t]

\begin{center}

\centerline{\includegraphics[width=0.5\textwidth]{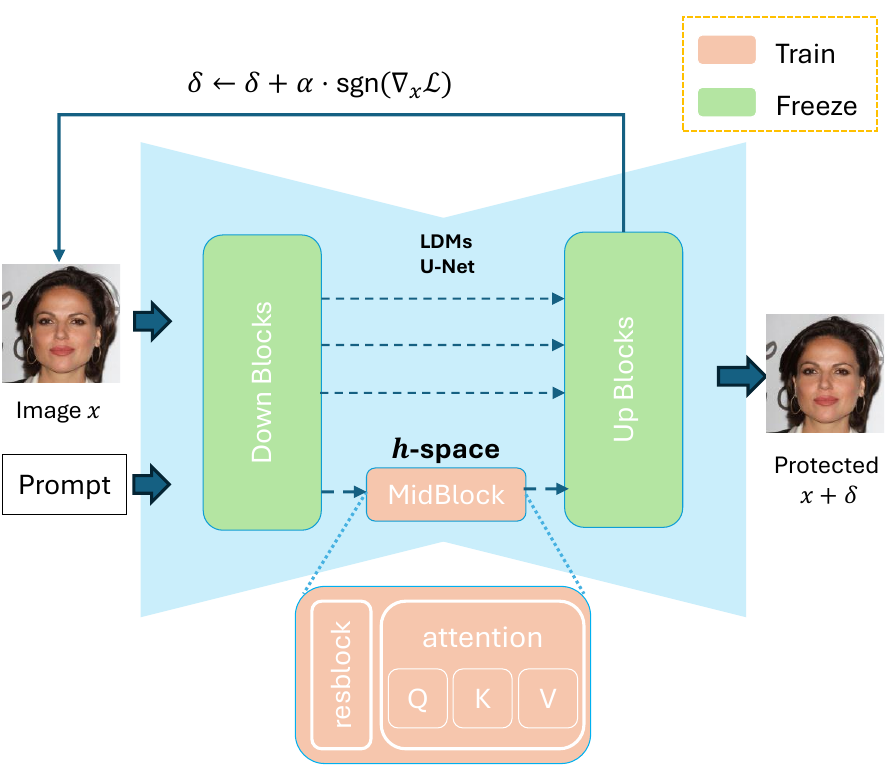}}
\vspace{-2mm}

\caption{Block diagram of {\ourmethod} method constructing the perturbed image $x+\delta$ based on the gradient of the loss function wrt to the \textit{h}-space. Only \textit{h}-space is optimized during training, while the rest of the model is kept frozen.}
\label{fig:haad-block-diagram}
\end{center}
\end{figure}

\subsection{{\ourmethod}: \textit{h}-space based Adversarial Attack on Diffusion Models}
The remarkable capability of diffusion models to generate user-specific content is rooted in their underlying semantic latent space \cite{kwon2023latentspace}. This semantic latent space (`\textit{h}-space'), represents the deep features from the middle-block of the U-Net architecture (see Fig. \ref{fig:haad-block-diagram}). Kwon et al \cite{kwon2023latentspace} found that the \textit{h}-space exhibits nice properties like homogeneity, linearity, robustness, and consistency across timesteps, similar to the latent space in GANs.  By exploring the properties of this space, researchers have been able to discover meaningful semantic directions that have application to image editing (manipulating facial appearance) \cite{haas2024fg,li2024cvpr,dalva2024cvpr,sparseAE2024neurips}and image-to-image translation \cite{wu2023iccv}. 

Inspired by the strong properties of the \textit{h}-space, we devise an adversarial attack leveraging this rich semantic feature (illustrated in Figure \ref{fig:haad-block-diagram}). We use the gradient of the reconstruction loss ($\mathcal{L}_{\text{LDM}}$) restricted to the \textit{h}-space to construct the adversarial perturbation using the iterative PGD method. 

Let $W_{h}$ represent the features (weights) of the \textit{h}-space, and $\delta$ denote the adversarial perturbation added to disrupt the semantic structure of the \textit{h}-space. The perturbation is computed iteratively using the Projected Gradient Descent(PGD) method approximately formulated as:
\[
x^{t+1} = \Pi_{\eta}\left(x^{t} + \alpha \cdot \text{sign} \left( \nabla_{x^t} \mathcal{L}_{\text{LDM}}(x^t, W^t_{h}) \right)\right),
\]
where: $\mathcal{L}_{\text{LDM}}$ is the reconstruction loss (Eq \ref{eq:loss}), $\Pi_{\eta}(\cdot)$ is the projection operator ensuring that $\|\delta\|_{\infty} \leq \eta$, where $\eta$ is the noise budget, $\alpha$ is the step size for the gradient update, $t$ represents the current iteration. $W^t_{h}$ indicates that the gradients were calculated with the updated \textit{h}-space after t step.

Following, we explain the reasoning behind training the model simultaneously while constructing the attack. Adversarial attacks can be constructed either to the full model or selectively to specific components. Attacks that are integrated into the training loop—such as those embedded within fine-tuning processes—interact with the model’s learning dynamics. These methods can adaptively influence the model’s parameter updates, leading to more effective adversarial perturbations. 
The same strategy was also employed by CAAT. In {\ourmethod}, we only update the parameters corresponding to the \textit{h}-space using the default optimization setting for the Stable Diffusion. By restricting the update to the \textit{h}-space, we ensure that the learned semantic features are directly affected by the adversarial perturbation while keeping the rest of the model intact. This tightly integrated training strategy enables the perturbation to interfere more effectively with the internal learning process of diffusion models—resulting in robust, semantically aligned protection that are highly transferable across personalization settings.

During training, we jointly perform two complementary operations within each optimization step. First, we compute the loss $\mathcal{L}_{\text{LDM}}$ used in latent diffusion models, and then apply a PGD step to determine a noise vector that maximizes the disruption of the reconstruction objective when injected into the image. This perturbation is added to the training images before the next iteration, encouraging the model to learn under adverse semantic conditions. This is repeated operation leads to a stronger protection.

Algorithm \ref{alg:haad}, presents the pseudocode of our untargeted adversarial attack to generate perturbations based on the \textit{h}-space as protection against unauthorized customization by a diffusion model. In our implementation, we freeze the entire diffusion model, only keeping the \textit{h}-space trainable and iteratively update its weights (similar to CAAT\cite{xu2024caat}), while constructing the attack perturbation to obtain strong protection. 

\begin{algorithm}[!t]
\caption{\ourmethod : \textit{h}-space based Adversarial Attack}\label{alg:haad}
\begin{algorithmic}[1]
\REQUIRE Image $x$, \textit{h}-space parameters $W_h$, step length $\alpha$, epochs $N$, budget $\eta$, learning rate $l$
\ENSURE Perturbed image $x'$
\STATE Initialize $\delta$
\FOR{$n=1$ $\to$ $N$}
    \STATE $\nabla_{W_h}, \nabla_x \gets \nabla_{LDM} (W_h, x+\delta)$
    \STATE $W_h \gets W_h - l \nabla_{W_h}$ $\triangleright$ \textit{h}-space weight updated.
    \STATE $\delta \gets \delta + \alpha$ sgn $(\nabla_{x})$ $\triangleright$ perturbation $\delta$ updated.
    \IF{$\delta > \eta$}     
        \STATE $\delta \gets {clip}(\delta, -\eta, \eta)$ $\triangleright$ ensures $||\delta||_{\infty} \leq \eta$   
    \ENDIF
    \STATE $x' \gets x+\delta$
\ENDFOR
\end{algorithmic}
\end{algorithm}

\noindent \textbf{HAAD-KV: An Improved Variant of HAAD.}
While HAAD introduces adversarial noise into the \textit{h}-space to disrupt high-level semantic representations, we further refine this strategy by narrowing the scope of perturbation to the key (K) and value (V) parameters within the cross-attention layers in the \textit{h}-space. This variant, referred to as HAAD-KV, leverages the fact that cross-attention plays a pivotal role in text-to-image diffusion models by creating an alignment between the input prompt and the generated visual content.

The motivation behind HAAD-KV stems from two key observations. First, during the personalization process, the cross-attention layers —especially the KV parameters — undergo substantial adaptation to capture new concepts. These components control how visual features attend to the semantic prompt over time, making them critical to preserving identity and fidelity in customized generation. Second, by focusing the attack on these parameters, we are able to maximize semantic disruption while updating only a minimal subset of parameters, thereby improving computational efficiency and reducing the perceptual footprint of the perturbation.

In practice, HAAD-KV operates similarly to HAAD in terms of optimization: we retain the PGD-based perturbation strategy guided by the reconstruction loss $\mathcal{L}_{\text{LDM}}$, but constrain the perturbation injection and gradient updates only to the KV parameters of the cross-attention layer within the \textit{h}-space. All other parts of the model remain frozen.

This approach not only enhances the interpretability of the attack — by isolating the exact mechanism through which text-image alignment is corrupted—but also leads to stronger overall attack performance. As demonstrated in our experiments (Section~\ref{sec:exp-results}), HAAD-KV consistently achieves greater degradation of personalized content than HAAD, despite requiring fewer parameter updates. These results highlight HAAD-KV as a highly effective and efficient protection strategy against unauthorized diffusion model customization. In {\Supp}, Sec. \ref{app:sec:theoretical}, we present a theoretical framework and supporting analysis that illustrates the relationship within the \textit{h}-space - specifically the key and value (KV) parameters in cross-attention layers - and the model’s semantic structure. We show that targeting these components can lead to significant semantic misalignment, thereby enhancing the strength of the protection.

\begin{figure*}
\begin{minipage}[t]{0.12\columnwidth}
    \centering
    \footnotesize
    \makebox{Clean}
\end{minipage}
\begin{minipage}[t]{0.12\columnwidth}
    \centering
    \footnotesize
    \makebox{No Attack}
\end{minipage}
\begin{minipage}[t]{0.12\columnwidth}
    \centering
    \footnotesize
    \makebox{AdvDM}
\end{minipage}
\begin{minipage}[t]{0.12\columnwidth}
    \centering
    \footnotesize
    \makebox{ACE}
\end{minipage}
\begin{minipage}[t]{0.12\columnwidth}
    \centering
    \footnotesize
    \makebox{ACE+}
\end{minipage}
\begin{minipage}[t]{0.12\columnwidth}
    \centering
    \footnotesize
    \makebox{CAAT}
\end{minipage}
\begin{minipage}[t]{0.12\columnwidth}
    \centering
    \footnotesize
    \makebox{{\ourmethod}}
\end{minipage}
\begin{minipage}[t]{0.12\columnwidth}
    \centering
    \footnotesize
    \makebox{{\ourmethod}-KV}
\end{minipage}

\centerline{\includegraphics[width=\textwidth]{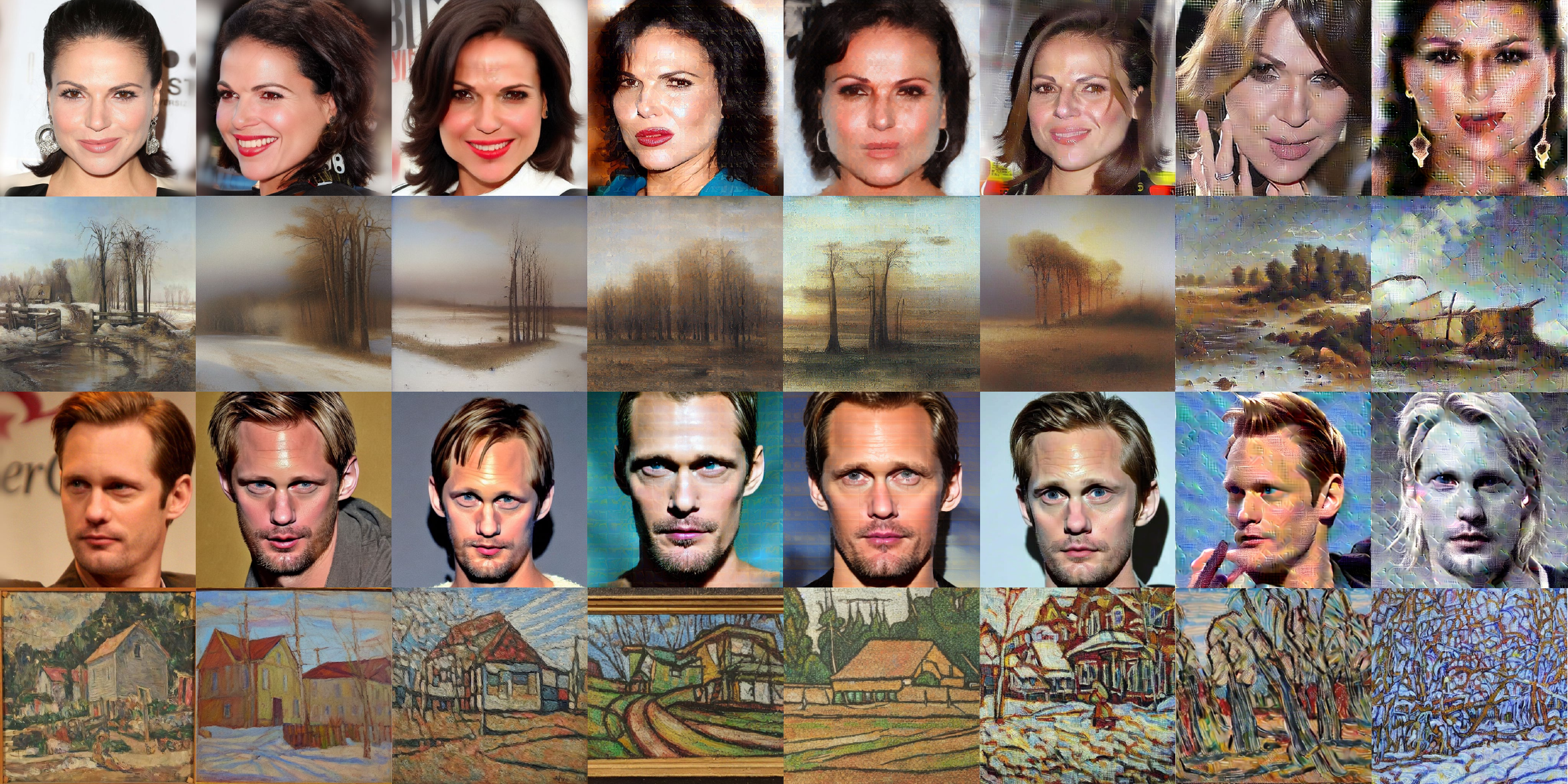}}

\caption{Images generated by popular diffusion models for customization: LoRA+Dreambooth (first two rows), Custom Diffusion (bottom two rows), using different adversarial protection methods with a noise budget of 4/255.}
\label{fig:main-comparison-visual}
\end{figure*}

\begin{table*}[htbp]
\caption{Comparison of our methods with other adversarial attack methods on LoRA+DB and CD. {\bf bold} indicates the best, while \underline{underline} indicates the second best. {\ourmethod}-KV achieves the best performance in 10 out of 12 metric-dataset combinations}
\centering
 \resizebox{\textwidth}{!}{
\renewcommand{\arraystretch}{0.9} 
\setlength{\tabcolsep}{5pt} 

\begin{tabular}{c|c c c c|c c c c|c c|c c}
\toprule
\multirow{9}{*}{} 
    & \multicolumn{8}{c|}{CelebA-HQ} 
    & \multicolumn{4}{c}{WikiArt} \\ \midrule
    & \multicolumn{4}{c|}{LoRA+DB} 
    & \multicolumn{4}{c|}{CD}  
    & \multicolumn{2}{c|}{LoRA+DB}
    & \multicolumn{2}{c}{CD} \\

            & CI $\uparrow$ & CS $\downarrow$ & FDFR $\uparrow$      & ISM $\downarrow$      & CI $\uparrow$ & CS $\downarrow$  & FDFR $\uparrow$     & ISM $\downarrow$   & CI $\uparrow$ & CS $\downarrow$ & CI $\uparrow$  & CS $\downarrow$      \\
            \midrule
No Attack   & 21.32    &   83.13    &0.005     &0.6904      &20.89      & 85.64    &0.005      &0.6907       & 34.62      & 64.51      & 34.22       & 66.36       \\ 
AdvDM       & 24.02    &   76.79    &0.005     &0.7633      &23.87      & 77.23    &0.015      &0.6721       & 35.32      & 61.12      & 36.45        & 61.85       \\ 
ACE         & 28.22    &    72.26   &0.015     &0.6081      &26.65      & 73.47    &0.055      &0.6065       & 37.01     & 58.69      &49.12        & 59.84       \\ 
ACE+        & 28.67    &  73.14   & 0.000     &0.6097      &28.20      & 74.01    &0.045       &0.6069       & 37.11      & 59.16      &\textbf{52.90}        & 60.13        \\ 
CAAT        & \underline{30.96}    &  72.17    &0.080      &0.5547      &\underline{29.31}       & \underline{73.26}      &\textbf{0.185}       &0.5763       & 37.52      &58.64       &51.17        & \underline{59.73}        \\
\midrule
{\bf \ourmethod}       & 29.10     &  \underline{72.06}    &\underline{0.085}       &\underline{0.5175}      &29.10       & 73.35     &\underline{0.150}       &\underline{0.5657}       & \underline{37.66}      & \underline{58.59}      &52.14        &  59.78    \\
{\bf{\ourmethod}-KV}      & \textbf{31.82} & \textbf{71.91} &\textbf{0.100}     &\textbf{0.5083}  &\textbf{29.52}& \textbf{72.98}    &\textbf{0.185}      &\textbf{0.5606}   & \textbf{37.88} & \textbf{58.26} &\underline{52.86}      & \textbf{59.52}        \\
\bottomrule

\end{tabular}
 }
\label{tab:main-comparison-qual}
\end{table*}

\section{Experimental Results} 
\label{sec:exp-results}

\begin{figure*}[!h]
\begin{center}
\centerline{\includegraphics[width=\textwidth]{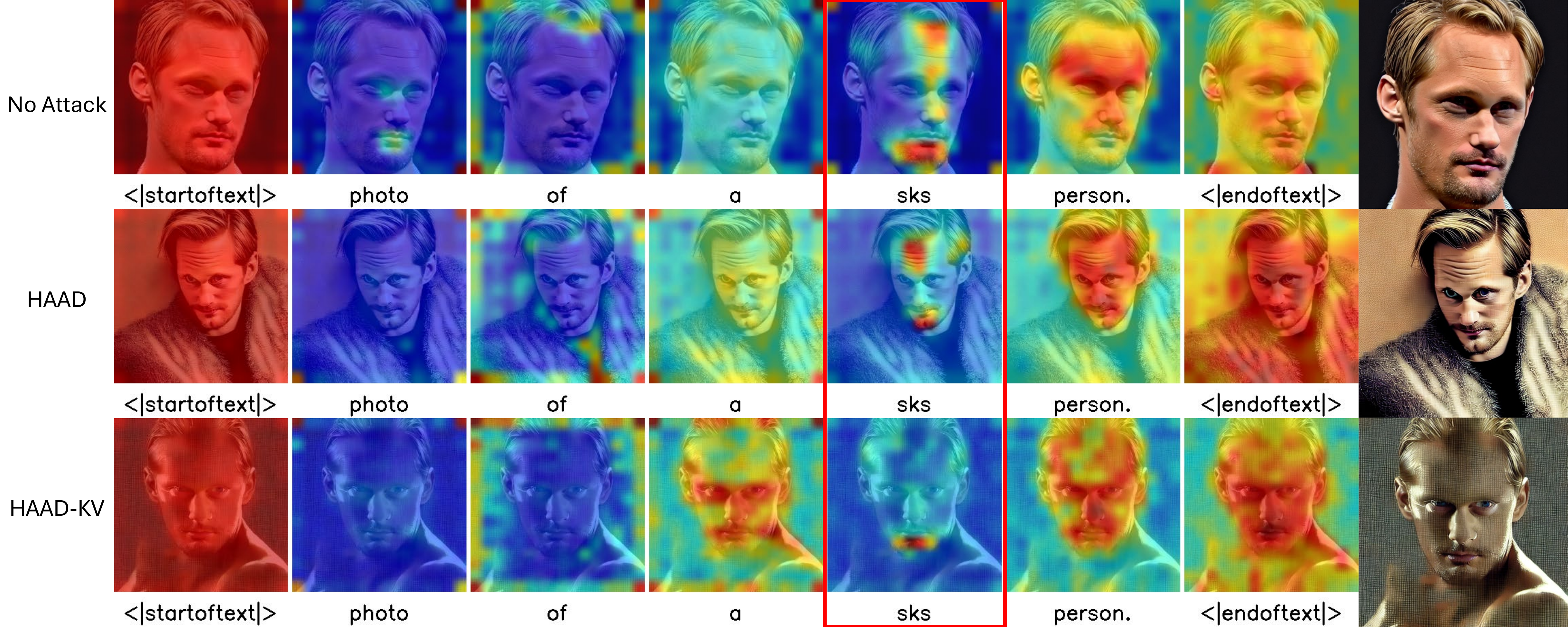}}
\caption{Visualization of the cross attention map of each token in the \textit{h}-space block during inference time: No Attack (first row), {\ourmethod} (middle row), and {\ourmethod}-KV (bottom row).}
\label{fig:discussion-cross-attn-map_main_paper}
\end{center}
\end{figure*}

\subsection{Experimental Setup}
\noindent\textbf{Datasets.} For the validation of our methods, we conducted experiments on two widely used datasets: CelebA-HQ \cite{karras2018celeba-hq} and WikiArt  \cite{saleh2015large} datasets. For CelebA-HQ, we select 200 images, corresponding to 10 persons (20 images per person), ensuring diversity in terms of gender, ethnicity and age. Similarly, for WikiArt, we choose 200 paintings from 10 artists (20 paintings per artist). 

\noindent\textbf{Customization Methods.}
We evaluate the effectiveness of our protection approach by applying it to several widely used customization pipelines. Specifically, we test on LoRA+DreamBooth \cite{HuSWALWWC22,ruiz2023db} (abbreviated as LoRA+DB) and Custom Diffusion (CD) \cite{kumari2023cd}, both of which represent state-of-the-art few-shot personalized concept learning approaches. These models capture different paradigms of personalization: LoRA+DB relies on parameter-efficient adaptation, while CD focuses on fine-tuning only the additional attention layers—providing a comprehensive evaluation ground for our methods. To ensure the reproducibility of our results, we use the publicly available code repositories for each method and follow their official hyper-parameter settings throughout our experiments.

\noindent\textbf{Comparison with SotA.} 
We compare the performance of HAAD and HAAD-KV with several representative adversarial based protection approaches, including AdvDM \cite{liang2023advdm}, ACE, ACE+ \cite{zheng2025targeted}, and CAAT \cite{xu2024caat}. These methods have been widely used in recent studies on protecting content from unauthorized diffusion model customization and provide a solid baseline for evaluating the effectiveness and efficiency of our proposed attacks.

\noindent\textbf{Details of the Adversarial Attacks.}
Our methods were computed based on SD1.5 as the reference model, as it is the most commonly deployed diffusion model. {\ourmethod} training focuses exclusively on the \textit{h}-space of Diffusion model \cite{sd2024}, using a batch size of 1, the learning rate of $1 \times 10^{-5}$ for 250 training steps. Mixed precision of bf16 is employed. The conditioning prompt used was "photo of a sks person/painting", where the `sks' user-specific token is initialized with `\textit{ktn}'. For the PGD attack, we set $\alpha=5 \times 10^{-3}$ (step size) and a noise budget of $\eta = 4/255$ in $\ell_{\infty}$-norm. All the experiments were run on a server with NVidia A40 GPU. For reproducibility, we provide all the settings in {\Supp}, Tables \ref{tab: hyperparameters} and \ref{tab:time-cost}.

Adversarial perturbation based anti-customization methods have studied their impact using different noise budgets ($\eta$). AdvDM was among the first works to report with a larger noise budget of $\eta = 32/255$ and similarly, CAAT with 0.1 ($\approx$ 25.5/255). Larger budgets tend to introduce perceptible noise that a human can easily detect visually, making them less suitable for real-world applications. Recently, ACE used a challenging setting with the noise budget of $\eta = 4/255$, where the noise in the perturbed image is indeed imperceptible. We choose the same strict setting of $\eta = 4/255$ in all our experiments to show that our approach works in the strictest condition. However, we perform a study to show the results of our approach with varying noise budgets (refer to {\Supp}, Figure \ref{app:fig:diff-eps} and Table \ref{app:tab:haad-kv-eps}).

\noindent\textbf{Evaluation metrics.} 
To assess the effectiveness of the evaluated protection methods, we adopted several quantitative metrics. CLIP Image-to-Image Similarity (CLIP-SIM) \cite{wang2023exploring}(shortened CS in tables), measures semantic similarity between generated and reference images based on CLIP embeddings. Lower scores  indicate stronger protection and semantic misalignment. We additionally use CLIP-IQA \cite{wang2023exploring} (shortened CI in tables), which evaluates perceptual image quality from a CLIP perspective. Higher CI scores reflect lower image quality and thus stronger disruption. For CelebA-HQ, we include two face-specific metrics: Face Detection Failure Rate (FDFR) \cite{deng2020retinaface}, where higher values indicate successful degradation of facial content, and Identity Score Matching (ISM) \cite{deng2019arcface}, where lower values indicate better obfuscation of personal identity.

\subsection{Qualitative Evaluation}
\begin{figure}[ht]
\begin{center}
\begin{minipage}[t]{0.10\textwidth}
    \centering
    \footnotesize
    \makebox{Data}
\end{minipage}
\begin{minipage}[t]{0.10\textwidth}
    \centering
    \footnotesize
    \makebox{No attack}
\end{minipage}
\begin{minipage}[t]{0.10\textwidth}
    \centering
    \footnotesize
    \makebox{AdvDM}
\end{minipage}
\begin{minipage}[t]{0.10\textwidth}
    \centering
    \footnotesize
    \makebox{CAAT}
\end{minipage}
\begin{minipage}[t]{0.10\textwidth}
    \centering
    \footnotesize
    \makebox{ACE}
\end{minipage}
\begin{minipage}[t]{0.10\textwidth}
    \centering
    \footnotesize
    \makebox{ACE+}
\end{minipage}
\begin{minipage}[t]{0.10\textwidth}
    \centering
    \footnotesize
    \makebox{{\ourmethod}}
\end{minipage}
\begin{minipage}[t]{0.10\textwidth}
    \centering
    \footnotesize
    \makebox{{\ourmethod}-KV}
\end{minipage}

\centerline{\includegraphics[width=0.9\textwidth]{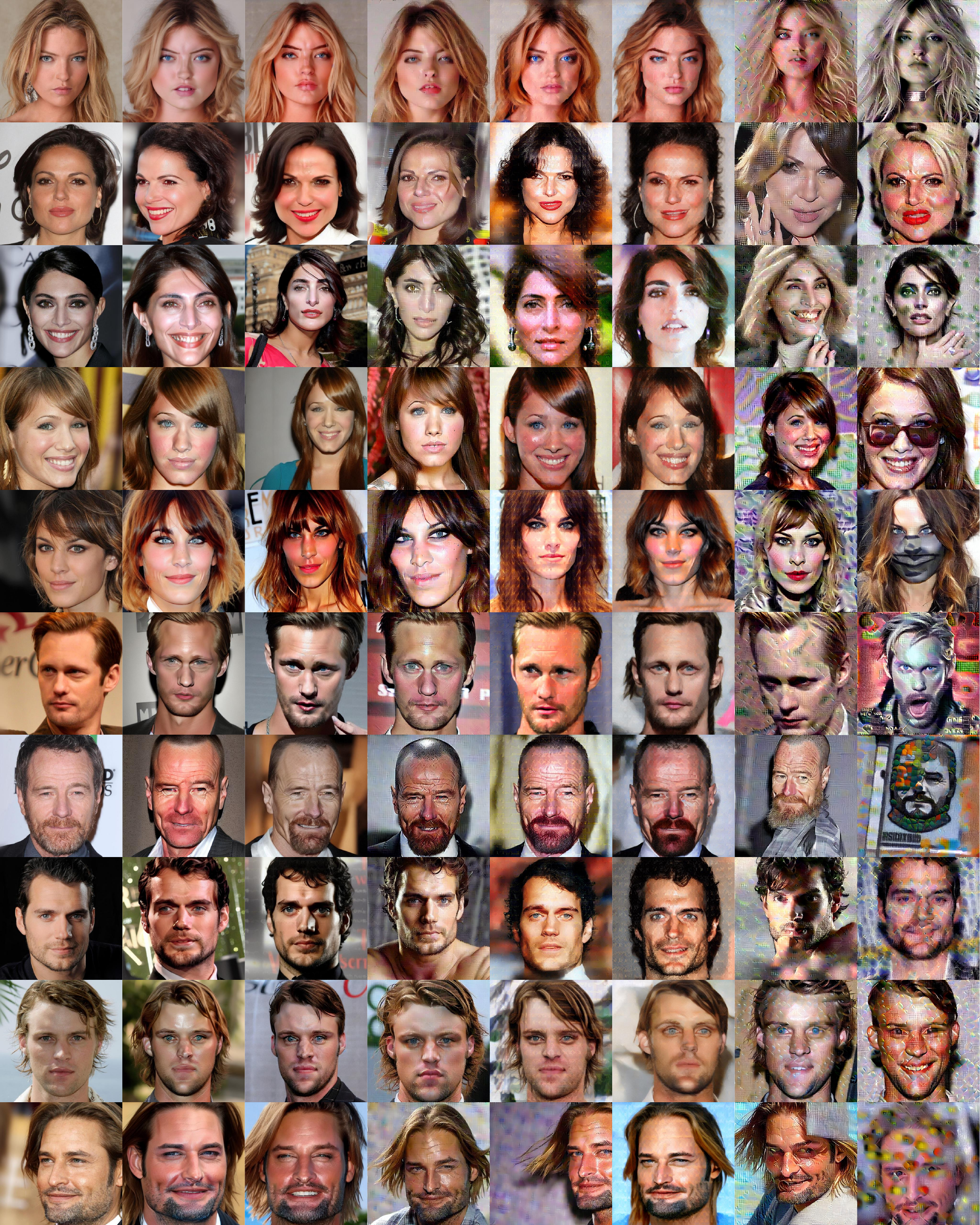}}
\caption{Comparison between different attack methods on CelebA-HQ dataset. The customization model used to generate the images is LoRA+DB.}
\label{app:fig:lora-face}
\end{center}
\vskip -0.2in
\end{figure}

\begin{figure*}[ht]
\begin{center}
\begin{minipage}[t]{0.11\textwidth}
    \centering
    \footnotesize
    \makebox{Data}
\end{minipage}
\begin{minipage}[t]{0.11\textwidth}
    \centering
    \footnotesize
    \makebox{No attack}
\end{minipage}
\begin{minipage}[t]{0.11\textwidth}
    \centering
    \footnotesize
    \makebox{AdvDM}
\end{minipage}
\begin{minipage}[t]{0.11\textwidth}
    \centering
    \footnotesize
    \makebox{CAAT}
\end{minipage}
\begin{minipage}[t]{0.11\textwidth}
    \centering
    \footnotesize
    \makebox{ACE}
\end{minipage}
\begin{minipage}[t]{0.11\textwidth}
    \centering
    \footnotesize
    \makebox{ACE+}
\end{minipage}
\begin{minipage}[t]{0.11\textwidth}
    \centering
    \footnotesize
    \makebox{{\ourmethod}}
\end{minipage}
\begin{minipage}[t]{0.11\textwidth}
    \centering
    \footnotesize
    \makebox{{\ourmethod}-KV}
\end{minipage}

\centerline{\includegraphics[width=0.9\textwidth]{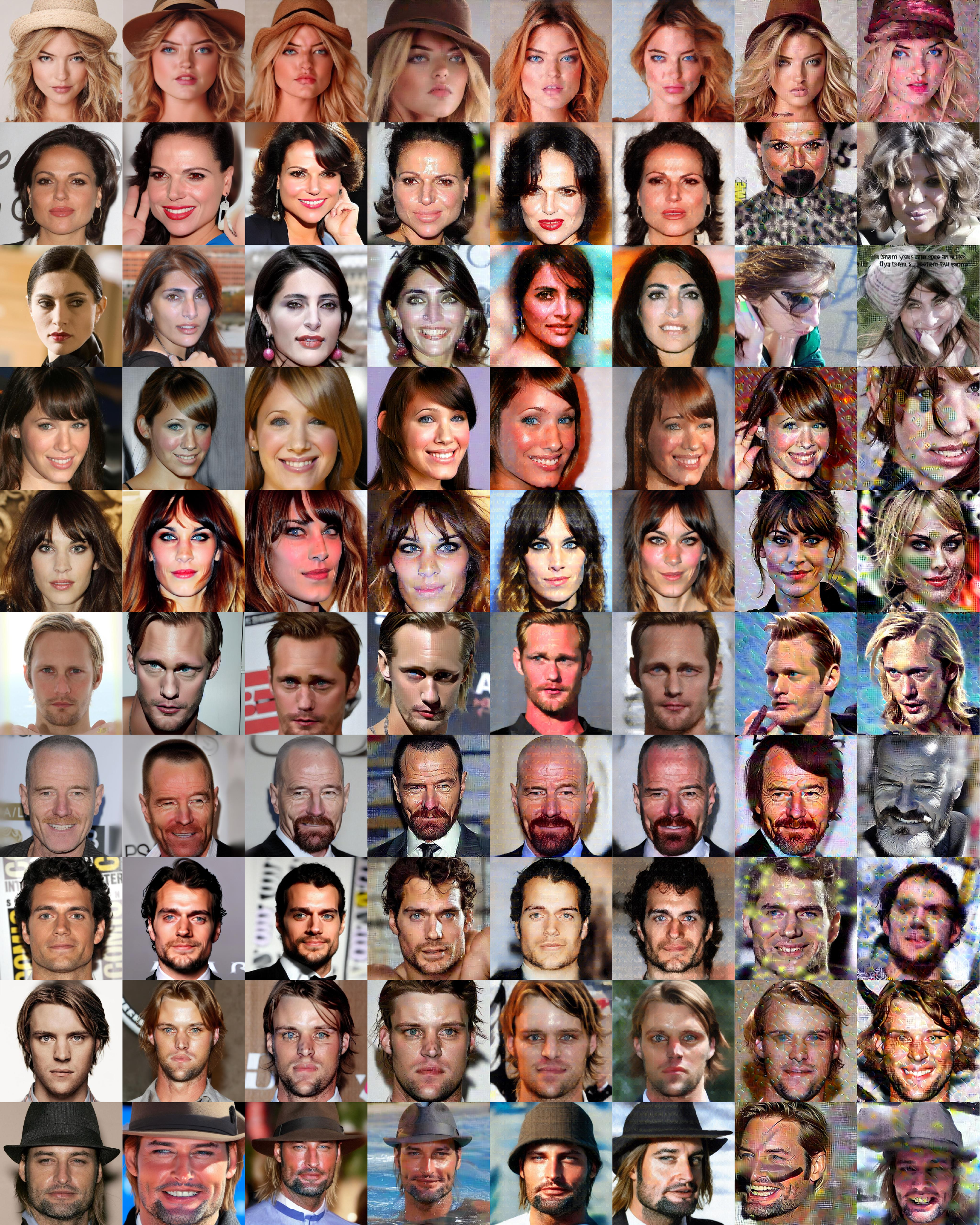}}
\caption{Comparison between different attack methods on CelebA-HQ dataset. The customization model used to generate the images is CD.}
\label{app:fig:cd-face}
\end{center}
\vskip -0.2in
\end{figure*}

\begin{figure*}[ht]
\begin{center}
\begin{minipage}[t]{0.11\textwidth}
    \centering
    \footnotesize
    \makebox{Data}
\end{minipage}
\begin{minipage}[t]{0.11\textwidth}
    \centering
    \footnotesize
    \makebox{No attack}
\end{minipage}
\begin{minipage}[t]{0.11\textwidth}
    \centering
    \footnotesize
    \makebox{AdvDM}
\end{minipage}
\begin{minipage}[t]{0.11\textwidth}
    \centering
    \footnotesize
    \makebox{CAAT}
\end{minipage}
\begin{minipage}[t]{0.11\textwidth}
    \centering
    \footnotesize
    \makebox{ACE}
\end{minipage}
\begin{minipage}[t]{0.11\textwidth}
    \centering
    \footnotesize
    \makebox{ACE+}
\end{minipage}
\begin{minipage}[t]{0.11\textwidth}
    \centering
    \footnotesize
    \makebox{{\ourmethod}}
\end{minipage}
\begin{minipage}[t]{0.11\textwidth}
    \centering
    \footnotesize
    \makebox{{\ourmethod}-KV}
\end{minipage}

\centerline{\includegraphics[width=0.9\textwidth]{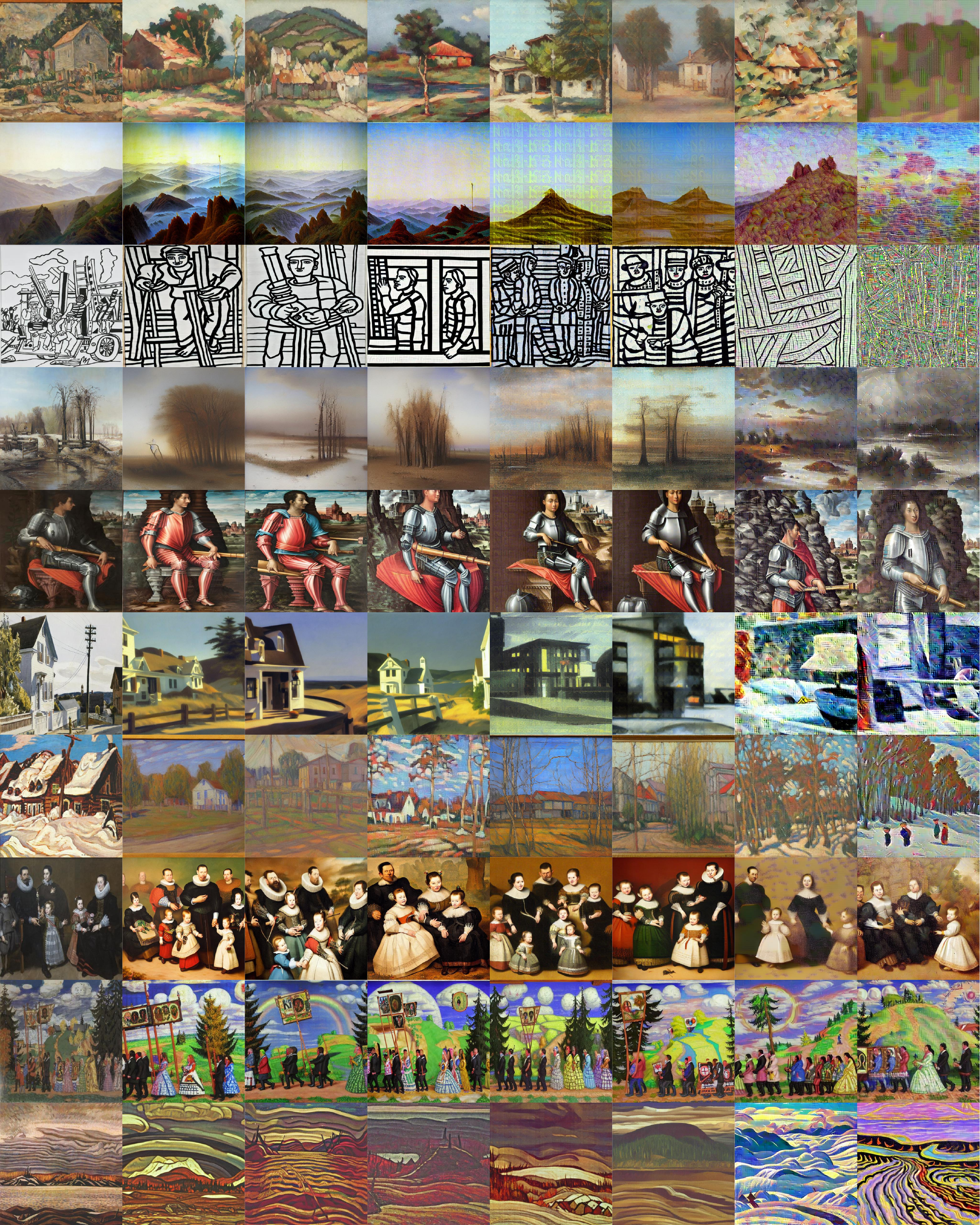}}

\caption{Comparison between different attack methods on WikiArt dataset. The customization model used to generate the images is LoRA+DB.}
\label{app:fig:lora-art}
\end{center}
\vskip -0.2in
\end{figure*}

\begin{figure*}[ht]
\begin{center}
\begin{minipage}[t]{0.11\textwidth}
    \centering
    \footnotesize
    \makebox{Data}
\end{minipage}
\begin{minipage}[t]{0.11\textwidth}
    \centering
    \footnotesize
    \makebox{No attack}
\end{minipage}
\begin{minipage}[t]{0.11\textwidth}
    \centering
    \footnotesize
    \makebox{AdvDM}
\end{minipage}
\begin{minipage}[t]{0.11\textwidth}
    \centering
    \footnotesize
    \makebox{CAAT}
\end{minipage}
\begin{minipage}[t]{0.11\textwidth}
    \centering
    \footnotesize
    \makebox{ACE}
\end{minipage}
\begin{minipage}[t]{0.11\textwidth}
    \centering
    \footnotesize
    \makebox{ACE+}
\end{minipage}
\begin{minipage}[t]{0.11\textwidth}
    \centering
    \footnotesize
    \makebox{{\ourmethod}}
\end{minipage}
\begin{minipage}[t]{0.11\textwidth}
    \centering
    \footnotesize
    \makebox{{\ourmethod}-KV}
\end{minipage}
\centerline{\includegraphics[width=0.9\textwidth]{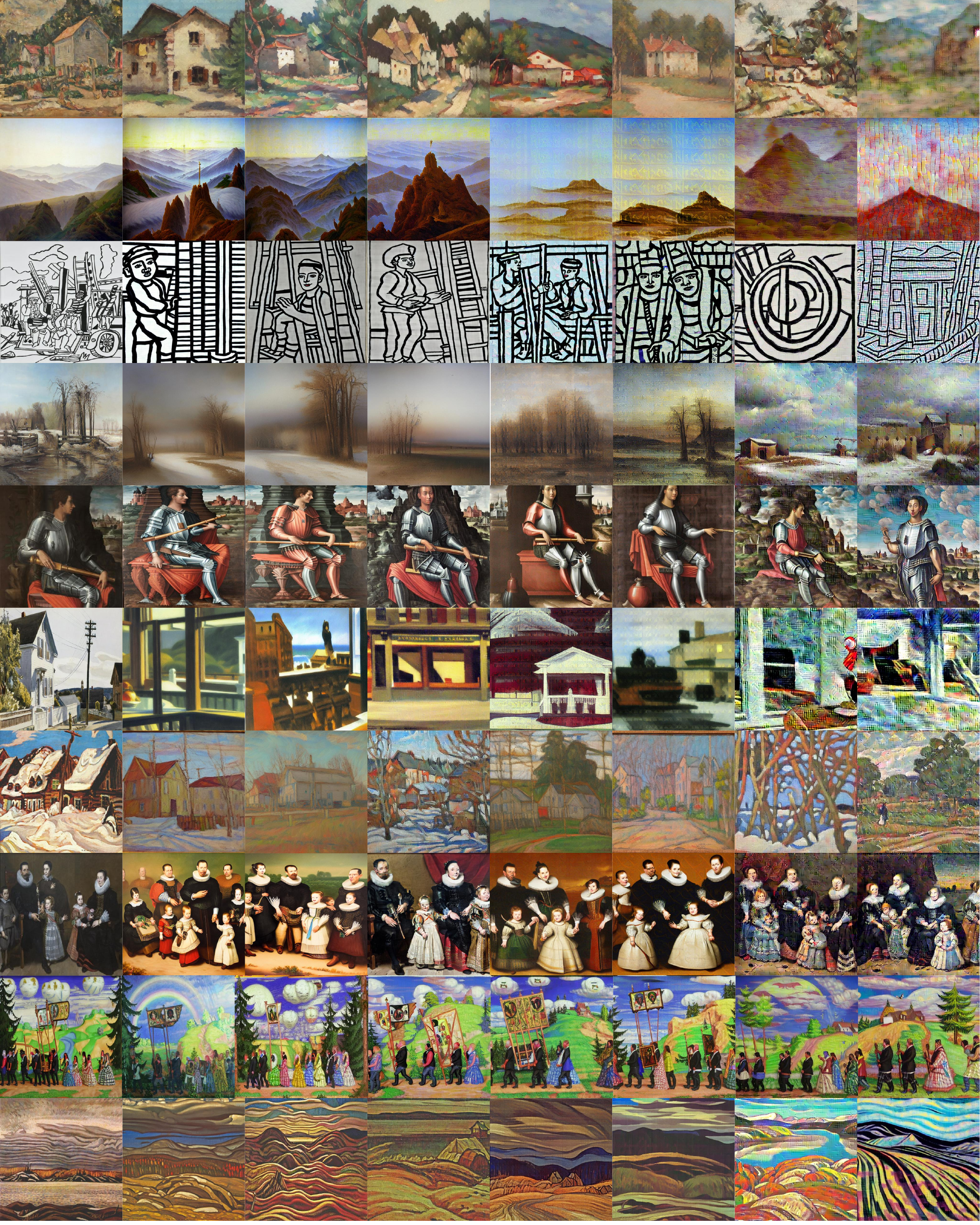}}
\caption{Comparison between different attack methods on WikiArt dataset. The customization model used to generate the images is CD.}
\label{app:fig:cd-art}
\end{center}
\vskip -0.2in
\end{figure*}

Figure \ref{fig:main-comparison-visual} shows that our attack significantly outperforms other approaches across LoRA+DB and CD. For LoRA+DB, our method introduces visible alterations both at the structural and semantic levels. For instance, we observe changes in facial attributes such as hair color, and in the case of art images, significant distortion of the main subject along with noticeable shifts in artistic style. For CD, the facial outputs undergo even more pronounced stylistic transformations, deviating substantially from the identity of the source individual. In artwork cases, the protected input leads to generation of entirely different visual compositions, often replacing the original subject with incoherent or unrelated patterns.
Additional visual comparisons for each model and content type can be found in {\Supp}, Figures \ref{app:fig:lora-face}, \ref{app:fig:cd-face}, \ref{app:fig:lora-art}, and \ref{app:fig:cd-art}.

\subsection{Quantitative Evaluation}

Table~\ref{tab:main-comparison-qual} summarizes the quantitative comparison between our methods (HAAD and HAAD-KV) and several state-of-the-art adversarial based protection techniques across two personalization settings: LoRA+DB and Custom Diffusion (CD). We evaluate performance using a range of perceptual, structural, and semantic metrics across both CelebA-HQ and WikiArt datasets.

We observe that HAAD achieves competitive performance, often outperforming AdvDM, ACE, and ACE+, and performing comparably to CAAT in several settings. Notably, HAAD achieves second-best results in most metrics and occasionally surpasses CAAT, particularly on CelebA-HQ in terms of ISM and FDFR.

However, once the attack is restricted to only the KV parameters in the cross-attention layer, i.e. the HAAD-KV variant, our method consistently achieves the best performance across nearly all metrics and setups. On the CelebA-HQ dataset, HAAD-KV achieves the best performance across all four metrics: it attains the highest CLIP-IQA (CI) and Face Detection Failure Rate (FDFR), as well as the lowest Identity Score Matching (ISM) and CLIP-SIM (CS), demonstrating its strong ability to degrade image quality, obscure identity, and disrupt semantic consistency. On the WikiArt dataset, HAAD-KV consistently ranks first or second in both CI and CS, confirming its effectiveness in protecting artistic content from unauthorized replication. When compared to CAAT, HAAD-KV achieves lower semantic similarity (CS), better perceptual degradation (CI), and more substantial structural distortion, indicating a clear advantage in both performance and efficiency.

These findings highlight the strength of our approach in disrupting both perceptual and semantic fidelity. By consistently outperforming existing methods across diverse models and datasets, HAAD-KV demonstrates its practical effectiveness as a generalizable protection against unauthorized customization.

\paragraph{User study:} We conducted an user study to evaluate how perceptually imperceptible the adversarial perturbations added to the images will be when used as a protection mechanism. The results are reported in {\Supp}, Sec. \ref{app:sec:user_study}.

\subsection{Protection Explainability and Parameter Efficiency}

To better understand the mechanism behind the effectiveness of our approach, we conduct a comparative analysis between HAAD and HAAD-KV from two complementary perspectives: semantic disruption in attention maps and training parameter efficiency.

We begin by visualizing the cross-attention maps corresponding to in the \textit{h}-space during inference (Figure~\ref{fig:discussion-cross-attn-map_main_paper}). We compare three settings: clean images (No attack), protected images with HAAD, and protected images with HAAD-KV. Red color highlights areas of the image which shows strong connection between text token and the generated image, while the blue color indicates the absence of such connection. 

In the `No attack' case, attention is tightly focused on meaningful facial regions (e.g., forehead and chin) and all non-relevant areas remain dark blue, indicating strong token-to-concept alignment.

With \ourmethod, attention becomes more erratic and partially misaligned. While central features still attract some focus, peripheral areas like the hairline and sides of the head begin to receive attention, showing that semantic grounding is weakening.

HAAD-KV results in further disruption. HAAD-KV leads to a complete loss of structure: the attention no longer clusters around any discernible region. Instead, it is diffusely spread across the background and unrelated parts of the image, with weak, scattered activations appearing in regions such as the neck, shoulders, or background textures. The map is characterized by widespread low-intensity coloring and an absence of concentrated attention hotspots, indicating that the model has effectively lost its ability to anchor the "sks" token to any consistent visual concept.

\begin{figure*}[ht]
\begin{center}
\centerline{\includegraphics[width=\textwidth]{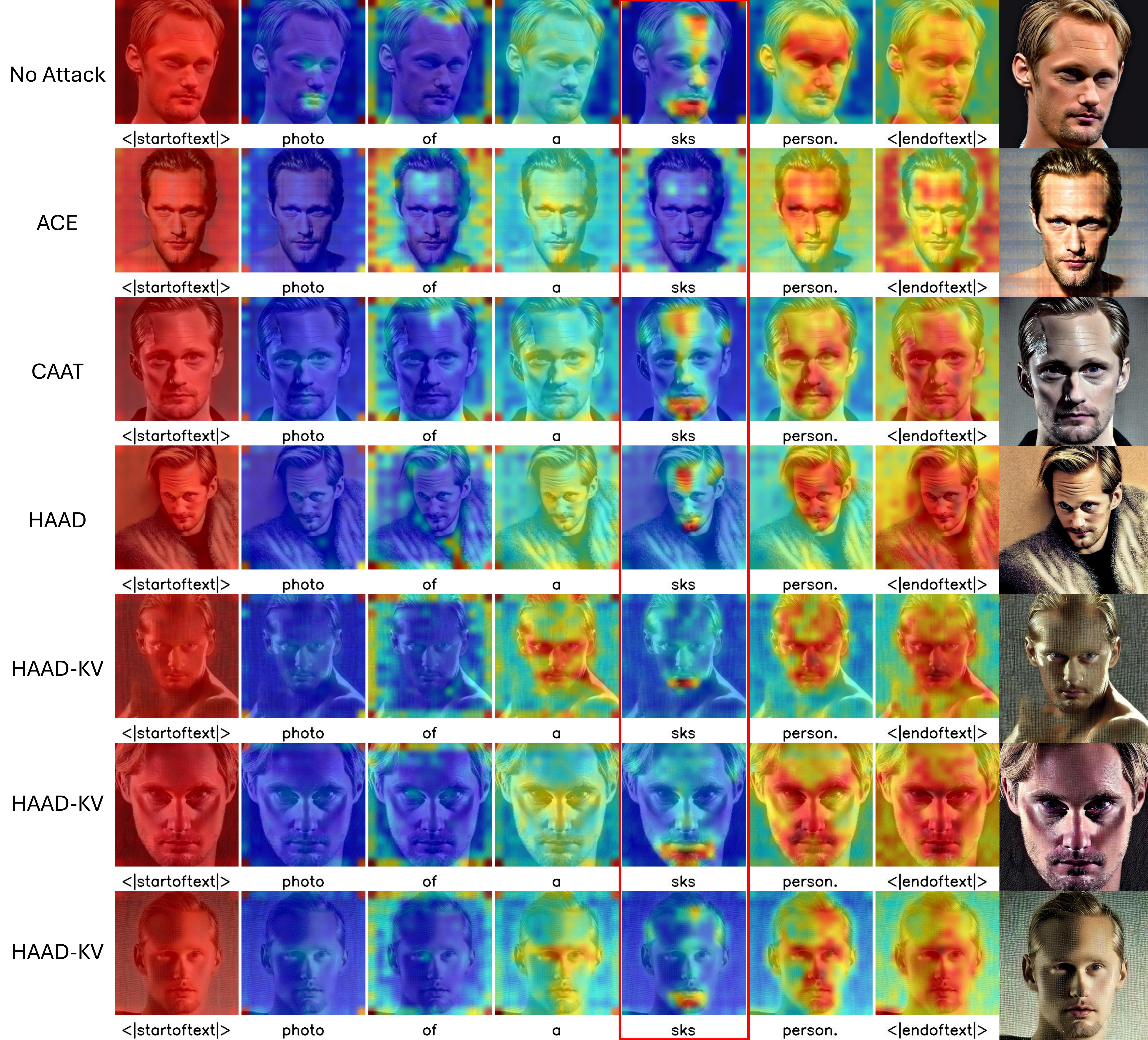}} 
\caption{Visualization of the cross attention map of each token during inference time for different adversarial attack methods on LoRA+DB with a sample of CelebA-HQ dataset.}
\label{app:fig:cross-attn-maps}
\end{center}
\vskip -0.2in
\end{figure*}

We also compare the parameter efficiency of different methods by reporting the number of trainable parameters updated while introducing the protection into the clean images (Table~\ref{tab:num_parameters}). HAAD-KV requires the fewest updated parameters—significantly fewer than all other baselines—while still achieving top performance. This highlights its lightweight nature and confirms that perturbing only the KV parameters in the cross-attention layer is sufficient to disrupt the personalization process effectively. A more complete visualization, involving other methods considered in this paper, is provided in {\Supp}, Figure \ref{app:fig:cross-attn-maps}.

\begin{table}[!h]
\caption{Number of updated parameters while introducing the protection into the clean images (in millions).}
\centering

\renewcommand{\arraystretch}{1.5} 
\setlength{\tabcolsep}{2pt} 
\begin{tabular}{c c c c c c}
\toprule
 AdvDM & ACE & ACE+ & CAAT & \ourmethod & \ourmethod-KV \\ \midrule
 859.52 & 123.06 & 123.06 & 19.17 & 97.03 & \textbf{5.24} \\ 
\bottomrule
\end{tabular}
\label{tab:num_parameters}

\end{table}

\subsection{Generalization and Robustness Analysis}

In this section, we study the impact of our methods on various operations: \begin{inparaenum} [(a)] \item purification methods (as studied in \cite{zheng2025targeted}) \item prompt invariance (as studied by \cite{vanle2023antidb}) \item image editing (example SDEdit\cite{MengHSSWZE22} as studied by ACE) and \item transferability to different models (as studied by ACE).
\end{inparaenum}

\subsubsection{Robustness to Purification Methods}

To evaluate the robustness of the protection introduced by adversarial perturbations, we follow prior work \cite{zheng2025targeted} to test whether common image pre-processing techniques can “purify” the perturbed images which could restore the original functionality of the diffusion model. Specifically, we apply a set of standard transformations including Gaussian noise ($\sigma=4, 8$), Gaussian blur (kernel size 3, 5), JPEG compression (quality 20, 70), resizing (including two setups: 2$\times$ up-scaling + recovering (2$\times$) and 0.5$\times$ down-scaling + recovering (0.5$\times$)), and super-resolution (SR) \cite{mustafa2019image} to the protected image.

We report CLIP-IQA (CI) scores under several representative purification settings in Table~\ref{tab:purification-comparison}, including Gaussian noise ($\sigma=8$), Gaussian blur (kernel size 5), JPEG ($Q=70$), resizing with $0.5\times$ and SR. A higher CI score indicates stronger protection. HAAD-KV consistently achieves the best performance across all shown methods, while HAAD ranks second in most cases. These results suggest that both variants are robust against common purification techniques. For a more complete analysis, refer to {\Supp}, Table \ref{app:tab:purfication} and Figure \ref{app:fig:purification}.

\begin{table}[htbp]
\caption{CLIP-IQA scores under selected purification settings.}
\centering

\renewcommand{\arraystretch}{0.9} 
\setlength{\tabcolsep}{5pt} 

\begin{tabular}{c|c|c|c|c|c }
\toprule

Protection / Defense     & Gaussian & Gaussian & JPEG & Resizing & SR \\ & noise ($\sigma=8$) & blur (ker. 5) &  ($Q=70$)  & (0.5$\times$) \\ \midrule
AdvDM       & 20.91  & 29.62 & 21.94 & 22.02 & 33.23 \\ 
ACE         & 25.96  & 28.55 & 23.11 & 25.91 & 35.26 \\ 
ACE+        & 24.68  & 29.35 & 22.89 & 26.06 & 34.76 \\ 
CAAT        & 27.81  & \underline{33.33} & 26.05 & \underline{26.38} & 35.55 \\
\midrule
{\bf \ourmethod} & \underline{28.71}  & 33.12 & \underline{28.87} & 26.37 & \underline{35.71} \\
{\bf{\ourmethod}-KV} & \textbf{29.15}  & \textbf{33.61} & \textbf{29.26} & \textbf{27.88} & \textbf{36.84} \\
\bottomrule

\end{tabular}

\label{tab:purification-comparison}
\end{table}

\subsubsection{Prompt Invariant Protection}

To evaluate the impact of the robustness of our methods against different prompts, we conduct inference using a diverse set of prompts \cite{vanle2023antidb} that describe different contexts and poses. Here we aim to assess whether protection remains effective when the prompt is changed while keeping the protected image fixed. We perform this experiment with six different prompts: "a dslr portrait of sks person", "a photo of sks person looking at the mirror", "a photo of sks person sitting on a chair", "a photo of sks person sitting on the floor", "a photo of sks person wearing glasses", "a photo of sks person talking on the phone".

\begin{figure*}[ht]
\begin{center}

\begin{minipage}[t]{0.16\textwidth}
    \centering
    \footnotesize
    \makebox{AdvDM}
\end{minipage}
\begin{minipage}[t]{0.16\textwidth}
    \centering
    \footnotesize
    \makebox{ACE}
\end{minipage}
\begin{minipage}[t]{0.16\textwidth}
    \centering
    \footnotesize
    \makebox{ACE+}
\end{minipage}
\begin{minipage}[t]{0.16\textwidth}
    \centering
    \footnotesize
    \makebox{CAAT}
\end{minipage}
\begin{minipage}[t]{0.16\textwidth}
    \centering
    \footnotesize
    \makebox{{\ourmethod}}
\end{minipage}
\begin{minipage}[t]{0.16\textwidth}
    \centering
    \footnotesize
    \makebox{{\ourmethod}-KV}
\end{minipage}

\centerline{\includegraphics[width=\textwidth]{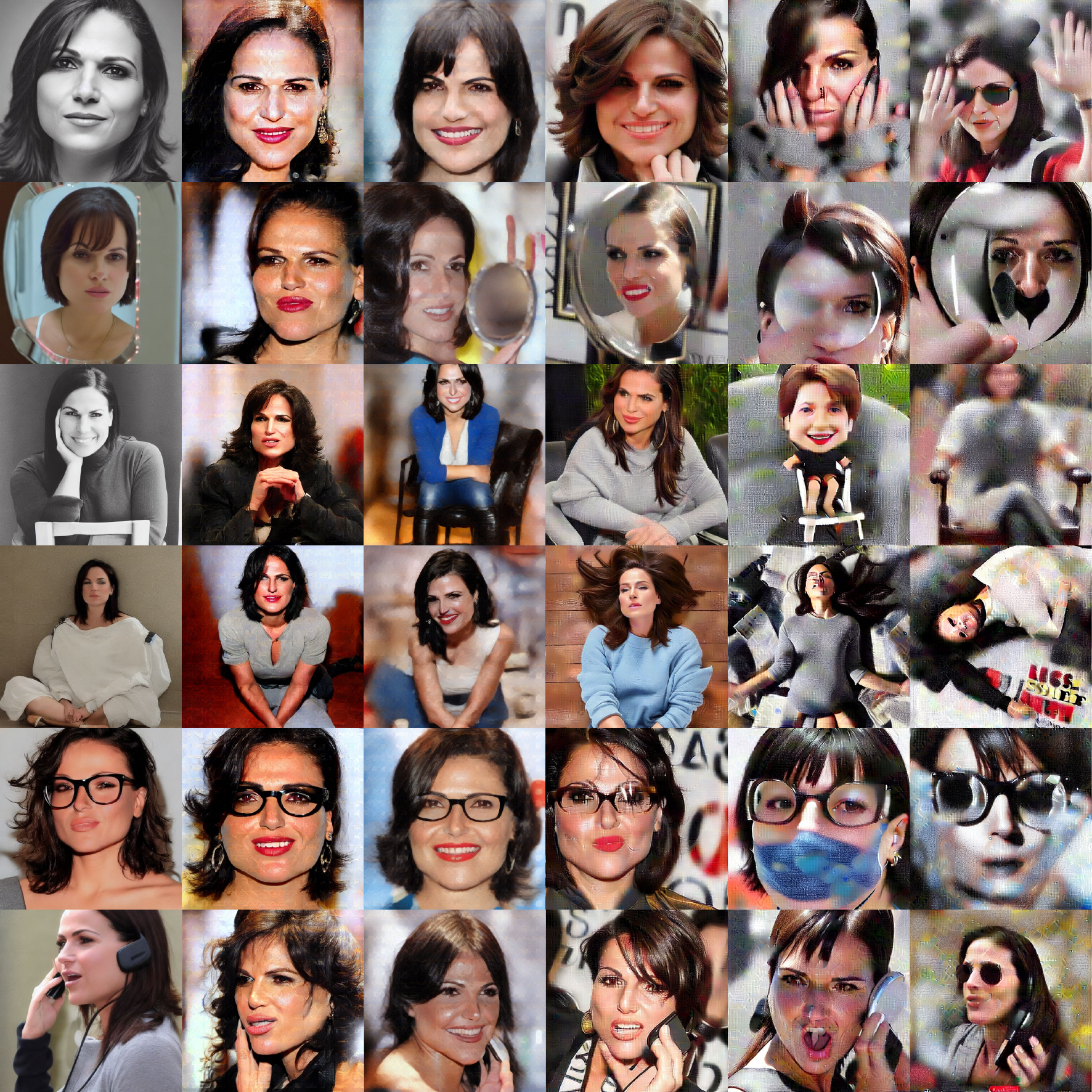}}

\caption{Results with different prompts during inference. (Row 1) "a dslr portrait of sks person", (Row 2) "a photo of sks person looking at the mirror", (Row 3) "a photo of sks person sitting on a chair", (Row 4) "a photo of sks person sitting on the floor", (Row 5) "a photo of sks person wearing glasses",(Row 6) "a photo of sks person talking on the phone".}
\label{app:fig:prompt}
\end{center}
\vskip -0.2in
\end{figure*}

Table \ref{tab:prompt-invariant} shows the quantitative results of "a dslr portrait of sks person". Across all metrics, We can observe HAAD and HAAD-KV achieve the best or second-best performances consistently, demonstrating the strong generalization to unseen contexts. Additional results for all six inference prompts are provided in {\Supp}, Table \ref{app:tab:prompt} and Figure \ref{app:fig:prompt}. These results confirm that our method generalizes effectively to diverse and unconstrained generation scenarios while remaining robust even when prompt semantics shift significantly.

\begin{table}[htbp]
\caption{Quantitative results with a different prompt "a dslr portrait of sks person" during inference.}
\centering

\renewcommand{\arraystretch}{0.9} 
\setlength{\tabcolsep}{5pt} 

\begin{tabular}{c|c c c c}
\toprule
\multirow{5}{*}{} 
    & \multicolumn{4}{c}{"a dslr portrait of sks person"} \\ \midrule 
            & CI $\uparrow$   & CS $\downarrow$ & FDFR $\uparrow$  & ISM $\downarrow$      \\
            \midrule
AdvDM       & 18.12    & 76.91     & 0.005      & 0.6366      \\ 
ACE         & 25.96    & 74.47     & 0.010      & 0.5921      \\ 
ACE+        & 25.77     & 75.10     & 0.007      & 0.5954      \\ 
CAAT        & \underline{29.48}     & 74.88     & 0.060      & 0.5917      \\
\midrule
{\bf \ourmethod}       & 29.37     & \underline{73.85}     & \underline{0.090}      & \underline{0.5899}    \\
{\bf {\ourmethod}-KV}      & \textbf{30.37}    & \textbf{72.09} & \textbf{0.115}    & \textbf{0.5682}      \\
\bottomrule

\end{tabular}

\label{tab:prompt-invariant}
\end{table}

\subsubsection{Protection against Image Editing}
\begin{figure*}[ht]
\begin{center}
\begin{minipage}[t]{0.11\textwidth}
    \centering
    \footnotesize
    \makebox{Data}
\end{minipage}
\begin{minipage}[t]{0.11\textwidth}
    \centering
    \footnotesize
    \makebox{No attack}
\end{minipage}
\begin{minipage}[t]{0.11\textwidth}
    \centering
    \footnotesize
    \makebox{AdvDM}
\end{minipage}
\begin{minipage}[t]{0.11\textwidth}
    \centering
    \footnotesize
    \makebox{CAAT}
\end{minipage}
\begin{minipage}[t]{0.11\textwidth}
    \centering
    \footnotesize
    \makebox{ACE}
\end{minipage}
\begin{minipage}[t]{0.11\textwidth}
    \centering
    \footnotesize
    \makebox{ACE+}
\end{minipage}
\begin{minipage}[t]{0.11\textwidth}
    \centering
    \footnotesize
    \makebox{{\ourmethod}}
\end{minipage}
\begin{minipage}[t]{0.11\textwidth}
    \centering
    \footnotesize
    \makebox{{\ourmethod}-KV}
\end{minipage}

\centerline{\includegraphics[width=0.9\textwidth]{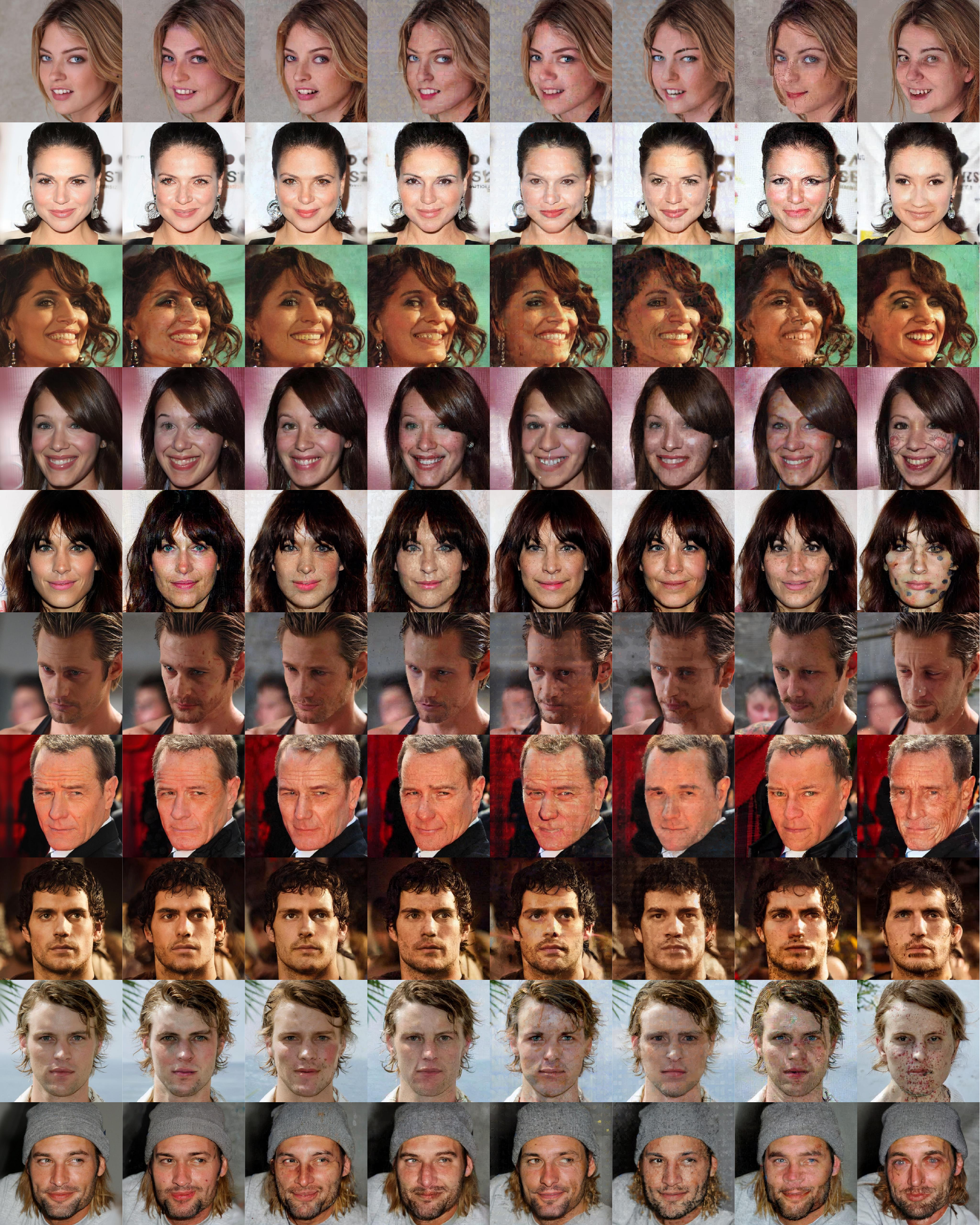}}
\caption{Comparison between different attack methods on CelebA-HQ dataset. The customization model used to generate the images is SDEdit.}
\label{app:fig:sdedit-face}
\end{center}
\vskip -0.2in
\end{figure*}

\begin{figure*}[ht]
\begin{center}
\begin{minipage}[t]{0.11\textwidth}
    \centering
    \footnotesize
    \makebox{Data}
\end{minipage}
\begin{minipage}[t]{0.11\textwidth}
    \centering
    \footnotesize
    \makebox{No attack}
\end{minipage}
\begin{minipage}[t]{0.11\textwidth}
    \centering
    \footnotesize
    \makebox{AdvDM}
\end{minipage}
\begin{minipage}[t]{0.11\textwidth}
    \centering
    \footnotesize
    \makebox{CAAT}
\end{minipage}
\begin{minipage}[t]{0.11\textwidth}
    \centering
    \footnotesize
    \makebox{ACE}
\end{minipage}
\begin{minipage}[t]{0.11\textwidth}
    \centering
    \footnotesize
    \makebox{ACE+}
\end{minipage}
\begin{minipage}[t]{0.11\textwidth}
    \centering
    \footnotesize
    \makebox{{\ourmethod}}
\end{minipage}
\begin{minipage}[t]{0.11\textwidth}
    \centering
    \footnotesize
    \makebox{{\ourmethod}-KV}
\end{minipage}
\centerline{\includegraphics[width=0.9\textwidth]{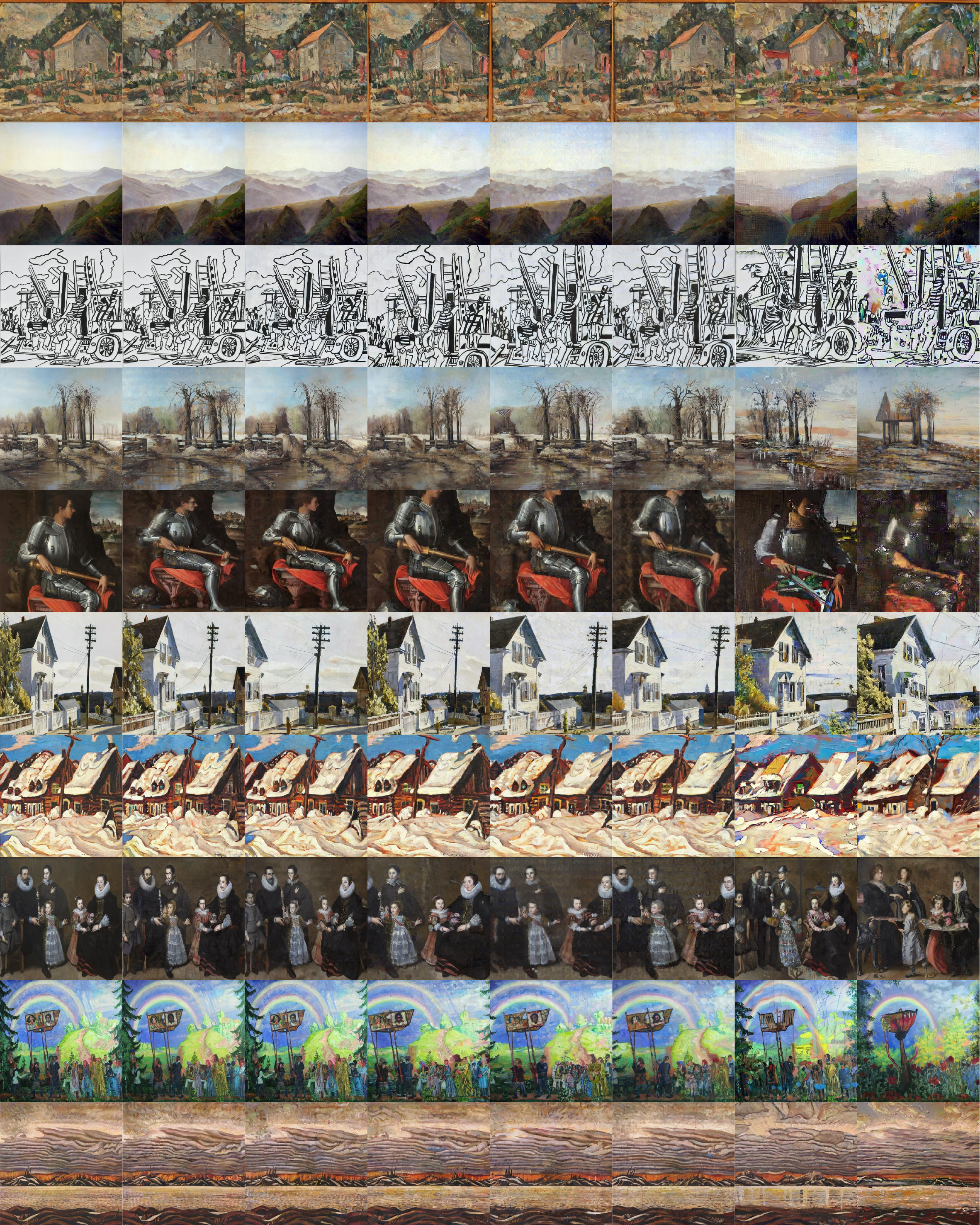}}
\caption{Comparison between different attack methods on WikiArt dataset. The customization model used to generate the images is SDEdit.}
\label{app:fig:sde-art}
\end{center}
\vskip -0.2in
\end{figure*}

While our primary goal was protection against concept-level customization, we also explore the potential of our method towards image editing through a preliminary study on SDEdit \cite{MengHSSWZE22}—a popular image-to-image editing framework based on diffusion models. Although SDEdit is not explicitly designed for concept learning, it can be used to make modifications to personal images with a prompt, raising practical concerns for misuse in identity editing. Table~\ref{tab:sdedit-results} reports performance using two evaluation metrics: Multi-Scale SSIM (MS) and CLIP-SIM (CS). Our method achieves the lowest scores across both metrics, suggesting that the perturbations remain partially effective even under this distinct editing paradigm. Notably, HAAD-KV achieves the best overall results on both datasets. More qualitative comparisons are shown in {\Supp}, Figures \ref{app:fig:sdedit-face}, \ref{app:fig:sde-art}. These results suggest that our method is not limited to text-to-image customization, but it may also offer a generalized protection against diffusion model-based image editing. While not our main focus, this preliminary study indicates that our attack strategy retains effectiveness even under structurally guided editing, laying a foundation for future research into editing-aware protections.

\begin{table}[htbp]
\caption{Comparison of our methods with other adversarial attack methods on SDEdit. {\bf bold} is the best, while \underline{underline} indicates the second best. {\ourmethod} and {\ourmethod}-KV achieves the best performance.}
\centering

\renewcommand{\arraystretch}{0.9} 
\setlength{\tabcolsep}{5pt} 

\begin{tabular}{c|c c|c c}
\toprule
\multirow{9}{*}{} 
    & \multicolumn{2}{c|}{CelebA-HQ} 
    & \multicolumn{2}{c}{WikiArt} \\ \midrule

            & MS $\downarrow$   & CS $\downarrow$ & MS $\downarrow$  & CS $\downarrow$      \\
            \midrule
No Attack   & 0.3637    & 79.26      & 0.1433      & 79.68      \\ 
AdvDM       & 0.3561    & 77.90     & 0.1421      & 77.38      \\ 
ACE         & 0.3393    & \textbf{75.53}     &  0.1411     & \textbf{75.39}      \\ 
ACE+        & 0.3366     & 76.27     & 0.1352      & 76.47      \\ 
CAAT        & 0.3488     & 77.16     & 0.1374      & 76.11      \\
\midrule
{\bf \ourmethod}       & \underline{0.3360}     & 76.10     & \textbf{0.1293}      & 75.85    \\
{\bf {\ourmethod}-KV}      & \textbf{0.3349}    & \underline{75.58} & \underline{0.1208}    & \underline{75.75}      \\
\bottomrule

\end{tabular}

\label{tab:sdedit-results}
\end{table}

\subsubsection{Transferability to different models}

\begin{table}[htbp]
\caption{CLIP IQA scores for Transferability of {\ourmethod} across different versions of SD.}
    \label{tab:transferability-sd-version-qual}
    \centering
    \renewcommand{\arraystretch}{0.9} 
    \setlength{\tabcolsep}{5pt} 
    \begin{tabular}{ c c c c}
            \hline
            Target      & SD1.4 & SD1.5  & SD2.1  \\ \midrule
            
            No Attack   & 18.89 & 21.32   & 19.18     
\\ 
            
            SD1.4       & 27.51 & 29.38   & 30.36     
\\ 
            
            SD1.5       & 29.32 & 29.10   & 30.55     
\\ 
            
            SD2.1       & 29.27 & 26.81   & 29.62     
\\ \hline

        \end{tabular}  
\end{table}

\begin{figure*}[h]
    \centering
    \begin{overpic}[width=0.95\textwidth]{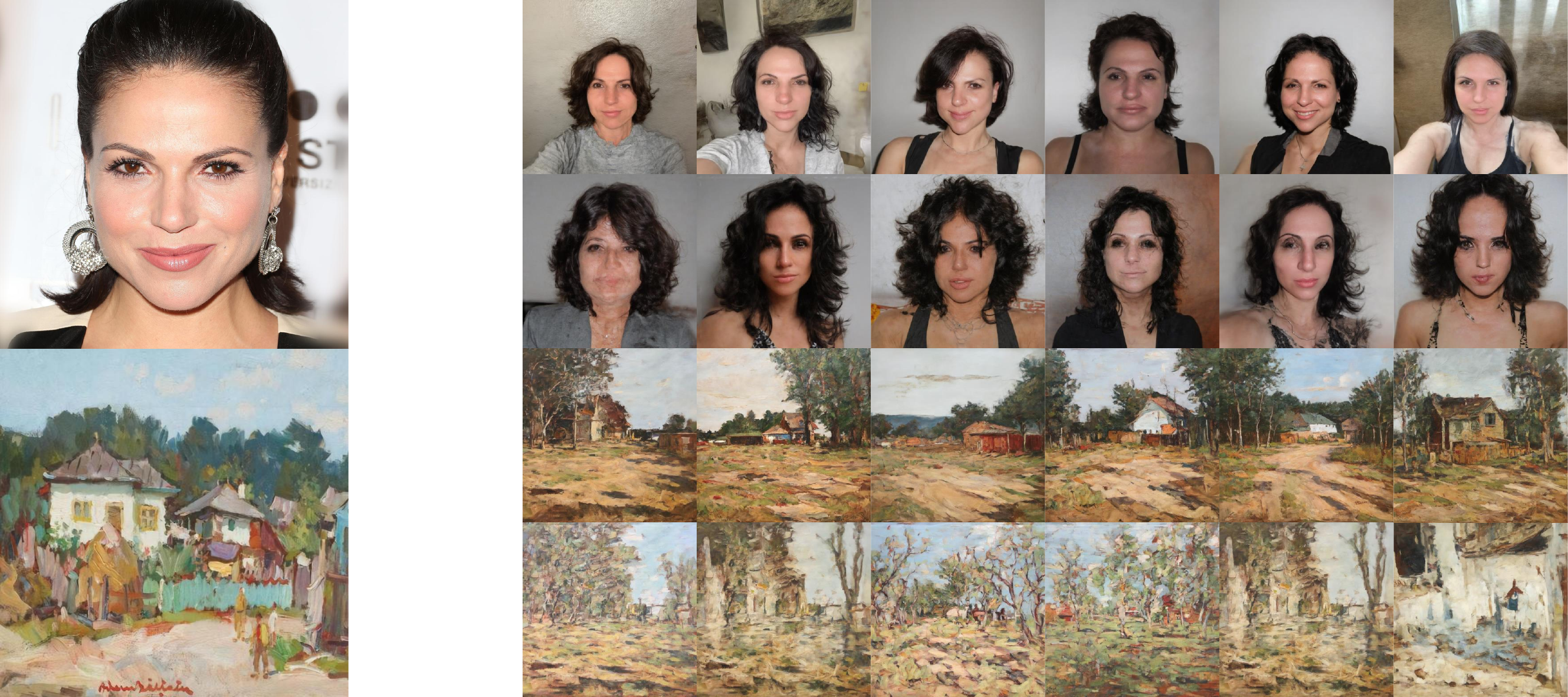} 

        \put(23.3,38){No Attack}
        \put(24,28){~\ourmethod}
        \put(23.3,16){No Attack}
        \put(24,5){~\ourmethod}
    \end{overpic}
    \caption{SD1.5 is used as the `Attacker' and SD3 as the `Target' model (used to generate the shown images). The figure highlights perceptible distortions in facial features, especially the eyes, while art images undergo distinct stylistic alterations. 
    }
    \label{app:fig:haad-transferability-sd3}
\end{figure*}

Transferability is an essential aspect of any protection method, as real-world scenario misuse may occur across diverse diffusion model architectures and versions. If a protection is tightly coupled to a specific model, it can be easily circumvented by switching to a different backbone. To evaluate this, we assess the robustness of our method on LoRA+Dreambooth in Table~\ref{tab:transferability-sd-version-qual}. For this experiment, multiple versions of Stable Diffusion (i.e., v1.4, v1.5, v2.1) are used in turn to generate adversarial perturbations (“Attacker”) and then evaluated on all three models (“Target”), covering both forward and backward transfer. Across all settings, {\ourmethod} retains strong protection ability, with minimal performance drop when transferred across SD versions. This demonstrates that our attack is not tied to a specific version but generalizes well within the Stable Diffusion family. In addition, we conducted a preliminary study on Stable Diffusion 3 (SD3) \cite{esser2024scaling}, a recently released model based on DiT (Diffusion Transformer) architecture. Using perturbations generated on SD1.5, we apply them to SD3 and observe visual degradation in both facial identity and artistic structure (refer {\Supp}, Figure \ref{app:fig:haad-transferability-sd3}). While these results are not yet conclusive, they suggest that our method may have initial transferability potential beyond UNet-based backbones, laying the groundwork for further exploration. More visual results on SD 1.4–2.1 are provided in {\Supp}, Table \ref{app:tab:transferability-sd-version-visual}.

\section{Conclusion}

In this work, by exploiting the semantic structure of the latent space of the U-Net (\textit{h}-space), we introduced HAAD and its more efficient variant HAAD-KV, two simple yet effective adversarial strategies designed to protect against unauthorized few-shot image personalization by diffusion models. HAAD, by focusing on the whole \textit{h}-space, disrupts the alignment between user-specific tokens and visual concepts. HAAD-KV, the computationally efficient variant, finds the perturbation based only on the KV parameters of the cross-attention layer in \textit{h}-space, achieving stronger protection with fewer trainable parameters. Extensive experiments across diverse personalization frameworks (LoRA+DB, CD) and datasets (CelebA-HQ, WikiArt) showed that our approaches consistently outperform existing methods, both in semantic distortion and perceptual degradation. We also conducted comprehensive robustness studies under standard image purification transformations, varying prompts, and multiple Stable Diffusion versions (including image editing as an additional use-case), validating the generalizability and transferability of our methods. Preliminary results on SD3 and SDEdit suggested that our approach may extend to broader generative pipelines, opening new directions for editing-aware and architecture-agnostic defenses. Overall, HAAD and HAAD-KV demonstrated that leveraging latent semantics offers a promising and efficient pathway towards safeguarding personal content in generative models.

\section*{Acknowledgements}
Xide Xu acknowledges the Chinese Scholarship Council (CSC) grant No.202306310064. This work is supported by Grant PID2022-143257NB-I00 funded by MCIN/AEI/10.13039/501100011033 and by "ERDF A Way of Making Europa", the Departament de Recerca i Universitats from Generalitat de Catalunya with reference 2021SGR01499, and the Generalitat de Catalunya CERCA Program.

\bibliographystyle{unsrt}  

\begin{thebibliography}{10}

\bibitem{kour2014real}
George Kour and Raid Saabne.
\newblock Real-time segmentation of on-line handwritten arabic script.
\newblock In {\em Frontiers in Handwriting Recognition (ICFHR), 2014 14th
  International Conference on}, pages 417--422. IEEE, 2014.

\bibitem{kour2014fast}
George Kour and Raid Saabne.
\newblock Fast classification of handwritten on-line arabic characters.
\newblock In {\em Soft Computing and Pattern Recognition (SoCPaR), 2014 6th
  International Conference of}, pages 312--318. IEEE, 2014.

\bibitem{hadash2018estimate}
Guy Hadash, Einat Kermany, Boaz Carmeli, Ofer Lavi, George Kour, and Alon
  Jacovi.
\newblock Estimate and replace: A novel approach to integrating deep neural
  networks with existing applications.
\newblock {\em arXiv preprint arXiv:1804.09028}, 2018.

\bibitem{Alpher02}
FirstName Alpher.
\newblock Frobnication.
\newblock {\em IEEE TPAMI}, 12(1):234--778, 2002.

\bibitem{Alpher05}
FirstName Alpher and FirstName Gamow.
\newblock Can a computer frobnicate?
\newblock In {\em CVPR}, pages 234--778, 2005.

\bibitem{ECCV2022}
Shai Avidan, Gabriel Brostow, Moustapha Cissé, Giovanni~Maria Farinella, and
  Tal Hassner, editors.
\newblock {\em Computer Vision -- ECCV 2022}.
\newblock Springer, 2022.

\bibitem{ho2020ddpm}
Jonathan Ho, Ajay Jain, and Pieter Abbeel.
\newblock Denoising diffusion probabilistic models.
\newblock In {\em Proc. of NeurIPS}, 2020.

\bibitem{song2021ddim}
Stefano~Ermon Jiaming~Song, Chenlin~Meng.
\newblock Denoising diffusion implicit models.
\newblock In {\em Proc. of ICLR}, 2021.

\bibitem{song2021score}
Yang Song, Jascha Sohl-Dickstein, Diederik~P. Kingma, Abhishek Kumar, Stefano
  Ermon, and Ben Poole.
\newblock Score-based generative modeling through stochastic differential
  equations.
\newblock In {\em Proc. of ICLR}, 2021.

\bibitem{dhariwal2021neurips}
Prafulla Dhariwal and Alex Nichol.
\newblock Diffusion models beat gans on image synthesis.
\newblock In {\em Proc. of NeurIPS}, 2021.

\bibitem{song2021maxlikelihood}
Yang Song, Conor Durkan, Iain Murray, and Stefano Ermon.
\newblock Maximum likelihood training of score-based diffusion models.
\newblock In {\em Proc. of NeurIPS}, 2021.

\bibitem{saharia2022imagen}
Chitwan Saharia, William Chan, Saurabh Saxena, Lala Li, Jay Whang, Emily
  Denton, Seyed Kamyar~Seyed Ghasemipour, Burcu~Karagol Ayan, et~al.
\newblock Photorealistic text-to-image diffusion models with deep language
  understanding.
\newblock In {\em Proc. of NeurIPS}, 2022.

\bibitem{ramesh2022dalle2}
Aditya Ramesh, Prafulla Dhariwal, Alex Nichol, Casey Chu, and Mark Chen.
\newblock Hierarchical text-conditional image generation with clip latents.
\newblock {\em arXiv preprint arXiv:2204.06125}, 2022.

\bibitem{rombach2022ldm}
Robin Rombach, Andreas Blattmann, Dominik Lorenz, Patrick Esser, and Bj\"orn
  Ommer.
\newblock High-resolution image synthesis with latent diffusion models.
\newblock In {\em Proc. of CVPR}, pages 10684--10695, 2022.

\bibitem{sd2024}
StabilityAI.
\newblock Stable diffusion.
\newblock \url{https://huggingface.co/stabilityai}, 2024.
\newblock Hugging Face Model Hub.

\bibitem{chefer2023ae}
Hila Chefer, Yuval Alaluf, Yael Vinker, Lior Wolf, and Daniel Cohen-Or.
\newblock Attend-and-excite: Attention-based semantic guidance for
  text-to-image diffusion models.
\newblock {\em ACM Trans. on Graphics}, 42(4):1--10, 2023.

\bibitem{daam2023acl}
Raphael Tang, Linqing Liu, Akshat Pandey, Zhiying Jiang, Gefei Yang, Karun
  Kumar, Pontus Stenetorp, Jimmy Lin, , and Ferhan Ture.
\newblock What the daam: Interpreting stable diffusion using cross attention.
\newblock In {\em Proc. of ACL}, volume~1, pages 5644–--5659, 2023.

\bibitem{zhang2024eccv}
Yang Zhang, Teoh~Tze Tzun, Lim~Wei Hern, Tiviatis Sim, and Kenji Kawaguchi.
\newblock Enhancing semantic fidelity in text-to-image synthesis: Attention
  regulation in diffusion models.
\newblock In {\em Proc. of ECCV}, 2024.

\bibitem{balaji2023eDiff}
Yogesh Balaji, Seungjun Nah, Xun Huang, Arash Vahdat, Jiaming Song, Qinsheng
  Zhang, Karsten Kreis, Miika Aittala, Timo Aila, Samuli Laine, Bryan
  Catanzaro, Tero Karras, and Ming-Yu Liu.
\newblock ediff-i: Text-to-image diffusion models with ensemble of expert
  denoisers.
\newblock {\em arXiv preprint arXiv:2211.01324}, 2023.

\bibitem{hertz2023p2p}
Amir Hertz, Ron Mokady, Jay Tenenbaum, Kfir Aberman, Yael Pritch, and Daniel
  Cohen-Or.
\newblock Prompt-to-prompt image editing with cross attention control.
\newblock In {\em Proc. of ICLR}, 2023.

\bibitem{mokady2023nulltext}
Ron Mokady, Amir Hertz, Kfir Aberman, Yael Pritch, and Daniel Cohen-Or.
\newblock Null-text inversion for editing real images using guided diffusion
  models.
\newblock In {\em Proc. of CVPR}, pages 6038--6047, 2023.

\bibitem{saharia2022palette}
Chitwan Saharia, William Chan, Huiwen Chang, Chris~A. Lee, Jonathan Ho, Tim
  Salimans, David~J. Fleet, and Mohammad Norouzi.
\newblock Palette: Image-to-image diffusion models.
\newblock In {\em Prof. of SIGGRAPH}, 2022.

\bibitem{poole2023dreamfusion}
Ben Poole, Ajay Jain, Jonathan~T. Barron, and Ben Mildenhall.
\newblock Dreamfusion: Text-to-3d using 2d diffusion.
\newblock In {\em Proc. of ICLR}, 2023.

\bibitem{xu2023dream3d}
Jiale Xu, Xintao Wang, Weihao Cheng, Yan-Pei Cao, Ying Shan, Xiaohu Qie, and
  Shenghua Gao.
\newblock Dream3d: Zero-shot text-to-3d synthesis using 3d shape prior and
  text-to-image diffusion models.
\newblock In {\em Proc. of CVPR}, pages 20908--20918, 2023.

\bibitem{tang2023makeit3D}
Junshu Tang, Tengfei Wang, Bo~Zhang, Ting Zhang, Ran Yi, Lizhuang Ma, and Dong
  Chen.
\newblock Make-it-3d: High-fidelity 3d creation from a single image with
  diffusion prior.
\newblock In {\em Proc. of ICCV}, pages 22819--22829, 2023.

\bibitem{ho2022vdm}
Jonathan Ho, Tim Salimans, Alexey Gritsenko, William Chan, Mohammad Norouzi,
  and David~J. Fleet.
\newblock Video diffusion models.
\newblock {\em arXiv preprint arXiv:2204.03458}, 2022.

\bibitem{blattmann2023svd}
Andreas Blattmann, Tim Dockhorn, Sumith Kulal, Daniel Mendelevitch, Maciej
  Kilian, Dominik Lorenz, et~al.
\newblock Stable video diffusion: Scaling latent video diffusion models to
  large datasets.
\newblock {\em arXiv preprint arXiv:2311.15127}, 2023.

\bibitem{gupta2023video}
Agrim Gupta, Lijun Yu, Kihyuk Sohn, Xiuye Gu, Meera Hahn, Li~Fei-Fei, Irfan
  Essa, Lu~Jiang, and José Lezama.
\newblock Photorealistic video generation with diffusion models.
\newblock {\em arXiv preprint arXiv:2312.06662}, 2023.

\bibitem{wolleb2022miccai}
Julia Wolleb, Florentin Bieder, Robin Sandkühler, and Philippe~C. Cattin.
\newblock Diffusion models for medical anomaly detection.
\newblock In {\em Proc. of MICCAI}, pages 35--45, 2022.

\bibitem{masood2023deepfake}
Momina Masood, Marriam Nawaz, Khalid~Mahmood Malik, Ali Javed, and Aun Irtaza.
\newblock Deepfakes generation and detection: State-of-the-art, open
  challenges, countermeasures, and way forward.
\newblock {\em Applied Intelligence}, 53:3974--4026, 2023.

\bibitem{shan2023glaze}
Shawn Shan, Jenna Cryan, Emily Wenger, Haitao Zheng, Rana Hanocka, and Ben~Y.
  Zhao.
\newblock Glaze: Protecting artists from style mimicry by text-to-image models.
\newblock In {\em Proc. of USENIX Conference on Security Symposium}, pages
  2187--2204, 2023.

\bibitem{shan2024nightshade}
Shawn Shan, Wenxin Ding, Josephine Passananti, Stanley Wu, Haitao Zheng, and
  Ben~Y Zhao.
\newblock Nightshade: Prompt-specific poisoning attacks on text-to-image
  generative models.
\newblock In {\em 2024 IEEE Symposium on Security and Privacy (SP)}. IEEE,
  2024.

\bibitem{chesney2019deepfake}
Robert Chesney and Danielle Citron.
\newblock Deepfakes and the new disinformation war: The coming age of
  post-truth geopolitics.
\newblock {\em Foreign Affairs}, 98(1):147--155, 2019.

\bibitem{kwon2023latentspace}
Mingi Kwon, Jaeseok Jeong, and Youngjung Uh.
\newblock Diffusion models already have a semantic latent space.
\newblock In {\em Proc. of ICLR}, 2023.

\bibitem{wu2023iccv}
Chen~Henry Wu and Fernando de~la Torre.
\newblock A latent space of stochastic diffusion models for zero-shot image
  editing and guidance.
\newblock In {\em Proc. of ICCV}, pages 7378--7387, 2023.

\bibitem{haas2024fg}
Rene Haas, Inbar Huberman-Spiegelglas, Rotem Mulayoff, Stella Grasshof, Sami~S.
  Brandt, and Tomer Michaeli.
\newblock Discovering interpretable directions in the semantic latent space of
  diffusion models.
\newblock In {\em Proc. of Automatic Face and Gesture Recognition (FG)}, pages
  1--9, 2024.

\bibitem{li2024cvpr}
Hang Li, Chengzhi Shen, Philip Torr, Volker Tresp, and Jindong Gu.
\newblock Self-discovering interpretable diffusion latent directions for
  responsible text-to-image generation.
\newblock In {\em Proc. of CVPR}, pages 12006--12016, 2024.

\bibitem{sparseAE2024neurips}
Ayodeji Ijishakin, Ming~Liang Ang, Levente Baljer, Daniel Chee~Hian Tan,
  Hugo~Laurence Fry, Ahmed Abdulaal, Aengus Lynch, and James~H. Cole.
\newblock H-space sparse autoencoders.
\newblock In {\em NeurIPS, Wokshop on Safe Generative AI}, 2024.

\bibitem{dalva2024cvpr}
Yusuf Dalva and Pinar Yanardag.
\newblock Noiseclr: A contrastive learning approach for unsupervised discovery
  of interpretable directions in diffusion models.
\newblock In {\em Proc. of CVPR}, pages 24209--24218, 2024.

\bibitem{liu2024eccv}
Runtao Liu, Ashkan Khakzar, Jindong Gu, Qifeng Chen, Philip Torr, and Fabio
  Pizzati.
\newblock Latent guard: a safety framework for text-to-image generation.
\newblock In {\em Proc. of ECCV}, 2024.

\bibitem{zhang2025surveyadvattack}
Chenyu Zhang, Mingwang Hu, Wenhui Li, and Lanjun Wang.
\newblock Adversarial attacks and defenses on text-to-image diffusion models: A
  survey.
\newblock {\em Information Fusion}, 114(102701):1--15, 2025.

\bibitem{xu2024privacy}
Xide Xu, Muhammad~Atif Butt, Sandesh Kamath, and Bogdan Raducanu.
\newblock Privacy protection in personalized diffusion models via targeted
  cross-attention adversarial attack.
\newblock {\em arXiv preprint arXiv:2411.16437}, 2024.

\bibitem{bau2019iclr}
David Bau, Jun-Yan Zhu, Hendrik Strobelt, Bolei Zhou, Joshua~B. Tenenbaum,
  William~T. Freeman, and Antonio Torralba.
\newblock Gan dissection: Visualizing and understanding generative adversarial
  networks.
\newblock In {\em Proc. of ICLR}, 2019.

\bibitem{jia2022neurips}
Shuai Jia, Bangjie Yin, Taiping Yao, Shouhong Ding, Chunhua Shen, Xiaokang
  Yang, and Chao Ma.
\newblock Adv-attribute: Inconspicuous and transferable adversarial attack on
  face recognition.
\newblock In {\em Proc. of NeurIPS}, 2022.

\bibitem{yang2024ijcb}
Jing Yang, Runping Xi, Yingxin Lai, Xun Lin, and Zitong Yu.
\newblock Ddap: Dual-domain anti-personalization against text-to-image
  diffusion models.
\newblock In {\em Proc. of IEEE Int'l. Joint Conference on Biometrics (IJCB)},
  pages 1--10, 2024.

\bibitem{kumari2023cd}
Nupur Kumari, Bingliang Zhang, Richard Zhang, Eli Shechtman, and Jun-Yan Zhu.
\newblock Multi-concept customization of text-to-image diffusion.
\newblock In {\em Proc. of CVPR}, pages 1932--1941, 2023.

\bibitem{ruiz2023db}
Nataniel Ruiz, Yuanzhen Li, Varun Jampani, Yael Pritch, Michael Rubinstein, and
  Kfir Aberman.
\newblock Dreambooth: Fine tuning text-to-image diffusion models for
  subject-driven generation.
\newblock In {\em Proc. of CVPR}, pages 22500--22510, 2023.

\bibitem{gal2022ti}
Rinon Gal, Yuval Alaluf, Yuval Atzmon, Or~Patashnik, Amit~H. Bermano, Gal
  Chechik, and Daniel Cohen-Or.
\newblock An image is worth one word: Personalizing text-to-image generation
  using textual inversion.
\newblock In {\em Proc. of ICLR}, 2023.

\bibitem{han2023svdiff}
Ligong Han, Yinxiao Li, Han Zhang, Peyman Milanfar, Dimitris Metaxas, and Feng
  Yang.
\newblock Svdiff: Compact parameter space for diffusion fine-tuning.
\newblock In {\em Proc. of ICCV}, pages 7323--7334, 2023.

\bibitem{chen2024disenbooth}
Hong Chen, Yipeng Zhang, Simin Wu, Xin Wang, Xuguang Duan, Yuwei Zhou, and
  Wenwu Zhu.
\newblock Disenbooth: Identity-preserving disentangled tuning for
  subject-driven text-to-image generation.
\newblock In {\em Proc. of ICLR}, 2024.

\bibitem{shi2024instantbooth}
Jing Shi, Wei Xiong, Zhe Lin, and Hyun~Joon Jung.
\newblock Instantbooth: Personalized text-to-image generation without test-time
  finetuning.
\newblock In {\em Proc. of CVPR}, pages 8543--8552, 2024.

\bibitem{ruiz2024hyperdb}
Nataniel Ruiz, Yuanzhen Li, Varun Jampani, Wei Wei, Tingbo Hou, Yeal Pritch,
  Neal Wadhwa, Michael Rubinstein, and Kfir Aberman.
\newblock Hyperdreambooth: Hypernetworks for fast personalization of
  text-to-image models.
\newblock In {\em Proc. of CVPR}, pages 6527--6536, 2024.

\bibitem{vanle2023antidb}
Thanh Van~Le, Hao Phung, Thuan Hoang~Nguyen, Quan Dao, Ngoc Tran, and Anh Tran.
\newblock Anti-dreambooth: Protecting users from personalized text-to-image
  synthesis.
\newblock In {\em Proc. of ICCV}, pages 2116--2127, 2023.

\bibitem{liu2024metacloak}
Yixin Liu, Chenrui Fan, Yutong Dai, Xun Chen, Pan Zhou, and Lichao Sun.
\newblock Metacloak: Preventing unauthorized subject-driven text-to-image
  diffusion-based synthesis via meta-learning.
\newblock In {\em Proc. of CVPR}, pages 24219--24228, 2024.

\bibitem{goodfellow2015fgsm}
Ian~J. Goodfellow, Jonathon Shlens, and Christian Szegedy.
\newblock Explaining and harnessing adversarial examples.
\newblock In {\em Proc. of ICLR}, 2015.

\bibitem{madry2018pgd}
Aleksander Madry, Aleksandar Makelov, Ludwig Schmidt, Dimitris Tsipras, and
  Adrian Vladu.
\newblock Towards deep learning models resistant to adversarial attacks.
\newblock In {\em Proc. of ICLR}, 2018.

\bibitem{croce2020autopgd}
Francesco Croce and Matthias Hein.
\newblock Reliable evaluation of adversarial robustness with an ensemble of
  diverse parameter-free attacks.
\newblock In {\em Proc. of ICML}, volume 119, pages 2206--2216, 2020.

\bibitem{yu2021lafeat}
Yunrui Yu, Xitong Gao, and Cheng-Zhong Xu.
\newblock Lafeat: Piercing through adversarial defenses with latent features.
\newblock In {\em Proc. of CVPR}, pages 5735--5745, 2021.

\bibitem{rosenberg2021acm}
Ishai Rosenberg, Asaf Shabtai, Yuval Elovici, and Lior Rokach.
\newblock Adversarial machine learning attacks and defense methods in the cyber
  security domain.
\newblock {\em ACM Computing Surveys (CSUR)}, 54(5):1--36, 2021.

\bibitem{liang2023advdm}
Chumeng Liang, Xiaoyu Wu, Yang Hua, Jiaru Zhang, Yiming Xue, Tao Song, Zhengui
  Xue, Ruhui Ma, and Haibing Guan.
\newblock Adversarial example does good: Preventing painting imitation from
  diffusion models via adversarial examples.
\newblock In {\em Proc. of ICML}, volume 202, pages 20763--20786, 2023.

\bibitem{liang2023mist}
Chumeng Liang and Xiaoyu Wu.
\newblock Mist: Towards improved adversarial examples for diffusion models.
\newblock {\em arXiv preprint arXiv:2305.12683v1}, 2023.

\bibitem{salman2023icml}
Hadi Salman, Alaa Khaddaj, Guillaume Leclerc, Andrew Ilyas, and Aleksander
  Madry.
\newblock Raising the cost of malicious ai-powered image editing.
\newblock In {\em Proc. of ICML}, volume 202, pages 29894--29918, 2023.

\bibitem{xu2024caat}
Jingyao Xu, Yuetong Lu, Yandong Li, Siyang Lu, Dongdong Wang, and Xiang Wei.
\newblock Perturbing attention gives you more bang for the buck: Subtle imaging
  perturbations that efficiently fool customized diffusion models.
\newblock In {\em Proc. of CVPR}, pages 24534--24543, 2024.

\bibitem{xue2023diffpgd}
Haotian Xue, Alexandre Araujo, Bin Hu, and Yongxin Chen.
\newblock Diffusion-based adversarial sample generation for improved
  stealthiness and controllability.
\newblock In {\em Proc, of NeurIPS}, 2023.

\bibitem{xue2024iclr}
Haotian Xue, Chumeng Liang, Xiaoyu Wu, and Yongxin Chen.
\newblock Toward effective protection against diffusion-based mimicry through
  score distillation.
\newblock In {\em Proc. of ICLR}, 2024.

\bibitem{wang2024simac}
Feifei Wang, Zhentao Tan, Tianyi Wei, Yue Wu, and Qidong Huang.
\newblock Simac: A simple anti-customization method for protecting face privacy
  against text-to-image synthesis of diffusion models.
\newblock In {\em Proc. of CVPR}, pages 12047--12056, 2024.

\bibitem{liu2024disruptdiff}
Yisu Liu, Jinyang An, Wanqian Zhang, Dayan Wu, Jingzi Gu, Zheng Lin, and
  Weiping Wang.
\newblock Disrupting diffusion: Token-level attention erasure attack against
  diffusion-based customization.
\newblock {\em arXiv preprint arXiv:2405.20584}, 2024.

\bibitem{liu2015celeba}
Ziwei Liu, Ping Luo, Xiaogang Wang, and Xiaoou Tang.
\newblock Deep learning face attributes in the wild.
\newblock In {\em Proc. of ICCV}, pages 3730--3738, 2015.

\bibitem{karras2018celeba-hq}
Tero Karras, Timo Aila, Samuli Laine, and Jaakko Lehtinen.
\newblock Progressive growing of gans for improved quality, stability, and
  variation.
\newblock In {\em Proc. of ICLR}, 2018.

\bibitem{Rombach_2022_CVPR_stablediffusion}
Robin Rombach, Andreas Blattmann, Dominik Lorenz, Patrick Esser, and Bj\"orn
  Ommer.
\newblock High-resolution image synthesis with latent diffusion models.
\newblock In {\em Proceedings of the IEEE/CVF Conference on Computer Vision and
  Pattern Recognition (CVPR)}, pages 10684--10695, June 2022.

\bibitem{ronneberger2015unet}
Olaf Ronneberger, Philipp Fischer, and Thomas Brox.
\newblock U-net: Convolutional networks for biomedical image segmentation.
\newblock In {\em Medical Image Computing and Computer-Assisted
  Intervention--MICCAI 2015: 18th International Conference, Munich, Germany,
  October 5-9, 2015, Proceedings, Part III 18}, pages 234--241. Springer, 2015.

\bibitem{wang2003multiscale}
Zhou Wang, Eero~P Simoncelli, and Alan~C Bovik.
\newblock Multiscale structural similarity for image quality assessment.
\newblock In {\em The Thrity-Seventh Asilomar Conference on Signals, Systems \&
  Computers, 2003}, volume~2, pages 1398--1402. Ieee, 2003.

\bibitem{wang2023exploring}
Jianyi Wang, Kelvin~CK Chan, and Chen~Change Loy.
\newblock Exploring clip for assessing the look and feel of images.
\newblock In {\em Proceedings of the AAAI Conference on Artificial
  Intelligence}, volume~37, pages 2555--2563, 2023.

\bibitem{ruiz2022dreambooth}
Nataniel Ruiz, Yuanzhen Li, Varun Jampani, Yael Pritch, Michael Rubinstein, and
  Kfir Aberman.
\newblock Dreambooth: Fine tuning text-to-image diffusion models for
  subject-driven generation.
\newblock {\em CVPR}, 2023.

\bibitem{deng2020retinaface}
Jiankang Deng, Jia Guo, Evangelos Ververas, Irene Kotsia, and Stefanos
  Zafeiriou.
\newblock Retinaface: Single-shot multilevel face localisation in the wild.
\newblock In {\em Proc. of CVPR}, pages 5203--5212, 2020.

\bibitem{deng2019arcface}
Jiankang Deng, Jia Guo, Niannan Xue, and Stefanos Zafeiriou.
\newblock Arcface: Additive angular margin loss for deep face recognition.
\newblock In {\em Proc. of CVPR}, pages 4690--–4699, 2019.

\bibitem{heusel2017fid}
Martin Heusel, Hubert Ramsauer, Thomas Unterthiner, Bernhard Nessler, and Sepp
  Hochreiter.
\newblock Gans trained by a two time-scale update rule converge to a local nash
  equilibrium.
\newblock In {\em Proc. of NeurIPS}, 2017.

\bibitem{terhorst2020ser-fiq}
Philipp Terhorst, Jan~Niklas Kolf, Naser Damer, Florian Kirchbuchner, and Arjan
  Kuijper.
\newblock Ser-fiq: unsupervised estimation of face image quality based on
  stochastic embedding robustness.
\newblock In {\em Proc. of CVPR}, pages 5650--–5659, 2020.

\bibitem{schroff2015facenet}
Florian Schroff, Dmitry Kalenichenko, and James Philbin.
\newblock Facenet: A unified embedding for face recognition and clustering.
\newblock In {\em Proceedings of the IEEE conference on computer vision and
  pattern recognition}, pages 815--823, 2015.

\bibitem{PreechakulCWS22}
Konpat Preechakul, Nattanat Chatthee, Suttisak Wizadwongsa, and Supasorn
  Suwajanakorn.
\newblock Diffusion autoencoders: Toward a meaningful and decodable
  representation.
\newblock In {\em {IEEE/CVF} Conference on Computer Vision and Pattern
  Recognition, {CVPR} 2022, New Orleans, LA, USA, June 18-24, 2022}, pages
  10609--10619. {IEEE}, 2022.

\bibitem{HuSWALWWC22}
Edward~J. Hu, Yelong Shen, Phillip Wallis, Zeyuan Allen{-}Zhu, Yuanzhi Li,
  Shean Wang, Lu~Wang, and Weizhu Chen.
\newblock Lora: Low-rank adaptation of large language models.
\newblock In {\em TProc. of ICLR}, 2022.

\bibitem{zheng2025targeted}
Boyang Zheng, Chumeng Liang, and Xiaoyu Wu.
\newblock Targeted attack improves protection against unauthorized diffusion
  customization.
\newblock In {\em Proc. of ICLR}, 2025.

\bibitem{saleh2015large}
Babak Saleh and Ahmed Elgammal.
\newblock Large-scale classification of fine-art paintings: Learning the right
  metric on the right feature.
\newblock {\em arXiv preprint arXiv:1505.00855}, 2015.

\bibitem{MengHSSWZE22}
Chenlin Meng, Yutong He, Yang Song, Jiaming Song, Jiajun Wu, Jun{-}Yan Zhu, and
  Stefano Ermon.
\newblock Sdedit: Guided image synthesis and editing with stochastic
  differential equations.
\newblock In {\em Proc. of ICLR}, 2022.

\bibitem{zhao2024unlearnableexamplesdiffusionmodels}
Zhengyue Zhao, Jinhao Duan, Xing Hu, Kaidi Xu, Chenan Wang, Rui Zhang, Zidong
  Du, Qi~Guo, and Yunji Chen.
\newblock Unlearnable examples for diffusion models: Protect data from
  unauthorized exploitation, 2024.

\bibitem{esser2024scaling}
Patrick Esser, Sumith Kulal, Andreas Blattmann, Rahim Entezari, Jonas
  M{\"u}ller, Harry Saini, Yam Levi, Dominik Lorenz, Axel Sauer, Frederic
  Boesel, et~al.
\newblock Scaling rectified flow transformers for high-resolution image
  synthesis.
\newblock In {\em Forty-first International Conference on Machine Learning},
  2024.

\bibitem{honig2025adversarial}
Robert H{\"o}nig, Javier Rando, Nicholas Carlini, and Florian Tram{\`e}r.
\newblock Adversarial perturbations cannot reliably protect artists from
  generative {AI}.
\newblock In {\em Proc. of ICLR}, 2025.

\bibitem{an2024sd4privacy}
Jinyang An, Wanqian Zhang, Dayan Wu, Zheng Lin, Jingzi Gu, and Weiping Wang.
\newblock Sd4privacy: Exploiting stable diffusion for protecting facial
  privacy.
\newblock In {\em {IEEE} International Conference on Multimedia and Expo,
  {ICME} 2024, Niagara Falls, ON, Canada, July 15-19, 2024}, pages 1--6.
  {IEEE}, 2024.

\bibitem{mustafa2019image}
Aamir Mustafa, Salman~H Khan, Munawar Hayat, Jianbing Shen, and Ling Shao.
\newblock Image super-resolution as a defense against adversarial attacks.
\newblock {\em IEEE Transactions on Image Processing}, 29:1711--1724, 2019.

\bibitem{honig2024adversarial}
Robert H{\"o}nig, Javier Rando, Nicholas Carlini, and Florian Tram{\`e}r.
\newblock Adversarial perturbations cannot reliably protect artists from
  generative ai.
\newblock {\em arXiv preprint arXiv:2406.12027}, 2024.

\bibitem{cao2023impress}
Bochuan Cao, Changjiang Li, Ting Wang, Jinyuan Jia, Bo~Li, and Jinghui Chen.
\newblock {IMPRESS}: Evaluating the resilience of imperceptible perturbations
  against unauthorized data usage in diffusion-based generative {AI}.
\newblock In {\em Thirty-seventh Conference on Neural Information Processing
  Systems}, 2023.

\bibitem{thurstone1927scale}
Louis~Leon Thurstone.
\newblock A law of comparative judgment.
\newblock {\em Psychological Review}, 34:273--286, 1927.

\end{thebibliography}

\clearpage
\section{Supplementary Material}
\label{Supplementary}

The supplementary contains experimental details and additional results for the experiments in the main paper.

\subsection{Experimental Details : Hyperparameters} \label{app:exp-details-hyper}

\begin{table}[!h]
\caption{Hyperparameters for different attack methods. The parameters for all the methods are set to their default settings. Noise budget of $\eta=4/255$ was used in all our study.}
\centering
\renewcommand{\arraystretch}{1.2} 
\setlength{\tabcolsep}{2pt} 
\begin{tabular}{c c c c c}
\toprule
 Parameters & AdvDM & CAAT & ACE & \ourmethod \\ \midrule
 train steps & 1000 & 250 & 50 &  250 \\
 learning rate & $1 \times 10^{-4}$ & $1 \times 10^{-5}$ & $5 \times 10^{-6}$ & $1 \times 10^{-5}$   \\
 $\alpha$    & $2/255$ & $5 \times 10^{-3}$ & $5 \times 10^{-3}$ & $5 \times 10^{-3}$   \\
\bottomrule
\end{tabular}
\label{tab: hyperparameters}
\end{table}

\begin{table}[!h]
\caption{Hyperparameters for different custom diffusion models, with their default settings.}
\centering
\renewcommand{\arraystretch}{1.2} 
\setlength{\tabcolsep}{2pt} 
\begin{tabular}{c c c c c}
\toprule
 Parameters & LoRA+DB & CD \\ \midrule
 train steps & 1000 & 250  \\
 learning rate & $5 \times 10^{-5}$ & $1 \times 10^{-5}$    \\
 batchsize    & 1 & 2   \\
 LoRA rank  & 4 & - \\
 
\bottomrule
\end{tabular}
\label{tab:time-cost}
\end{table}

\subsection{Results with different noise budgets $\eta$ (HAAD vs HAAD-KV)} 
\label{app:different-eta-budget}

\begin{figure}[ht]
\centering
\begin{minipage}[t]{0.06\columnwidth}
    \centering
    \footnotesize
    \makebox{}
\end{minipage}
\begin{minipage}[t]{0.22\columnwidth}
    \centering
    \footnotesize
    \makebox{No Attack}
\end{minipage}
\begin{minipage}[t]{0.22\columnwidth}
    \centering
    \footnotesize
    \makebox{$\eta$=4/255}
\end{minipage}
\begin{minipage}[t]{0.22\columnwidth}
    \centering
    \footnotesize
    \makebox{$\eta$=8/255}
\end{minipage}
\begin{minipage}[t]{0.2\columnwidth}
    \centering
    \footnotesize
    \makebox{$\eta$=16/255}
\end{minipage}

\includegraphics[width=\columnwidth]{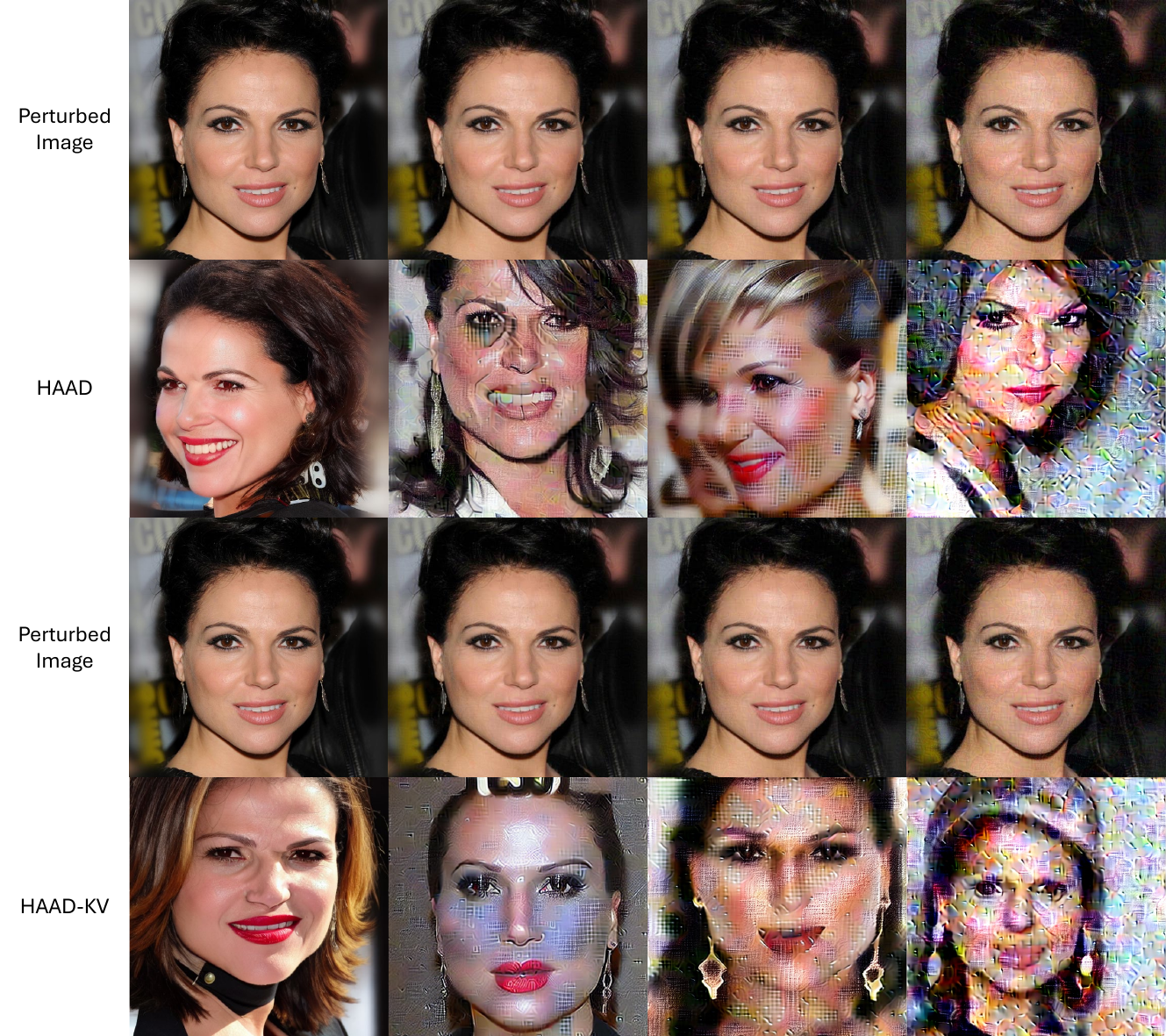}

\caption{Comparison between {\ourmethod} (first two rows) and {\ourmethod}-KV (second two rows) for varying noise budgets of \{0, 4/255, 8/255, 16/255\} (from left to right). We observe a consistent degradation of the generated images (using LoRA+DB as customization method).}
\label{app:fig:diff-eps}
\vskip -0.2in
\end{figure}

An increased noise budget ($\eta$) could result in perturbations that become more perceptible to the human eye. Figure \ref{app:fig:diff-eps} and Table \ref{app:tab:haad-kv-eps} show the effectiveness of attack with different noise budgets. At a low noise budget of 4/255 we find it hard to observe any visible change in the perturbation added to the image, hence, considered imperceptible to humans, but still creates noticeble changes to features in the generated image.  At noise budget of greater than 8/255, the perturbations start becoming more pronounced (observable with zoom in Figure \ref{app:fig:diff-eps}) and resulting in the creation of faint artificial patterns (e.g., grid-like stripes) in the generated images. At a higher noise budget of 16/255, the adversarial samples exhibit the largest change, resulting in the generated images to be almost completely unrecognizable. This noise level ensures the highest level of privacy protection, as the images become highly blurred and indistinct.

The results show that {\ourmethod}-KV offers better protection than {\ourmethod}. At the same time, even with a small noise budget of 4/255, our methods remains highly effective and are comparable to other methods operating at higher noise levels ($>8/255$). 


\begin{table}[htbp]

\caption{A quantitative comparison of {\ourmethod} and {\ourmethod}-KV on LoRA+DB with CelebA-HQ.}
\centering

\renewcommand{\arraystretch}{0.9} 
\setlength{\tabcolsep}{5pt} 

\begin{tabular}{c|c c|c c|c c}
\toprule
\multirow{9}{*}{} 
    & \multicolumn{2}{c|}{CI $\uparrow$} 
    & \multicolumn{2}{c|}{FDFR $\uparrow$}
    & \multicolumn{2}{c}{ISM $\downarrow$} \\ 
       & {\ourmethod}      & {\ourmethod}-KV    &  {\ourmethod}   & {\ourmethod}-KV     & {\ourmethod}   & {\ourmethod}-KV        \\ 
       \midrule
4/255  & 29.10  & 29.52 & 0.085 & 0.100   & 0.5175    & 0.5083    \\ 
8/255  & 37.70  & 40.01 & 0.110 & 0.180   & 0.4517    & 0.4384    \\ 
16/255 & 46.11  & 48.41 & 0.690 & 0.820   & 0.2905    & 0.2357    \\ 
\bottomrule

\end{tabular}

\label{app:tab:haad-kv-eps}
\end{table}

\clearpage
 \onecolumn
\subsection{Qualitative and quantitative results comparison of robustness between adversarial methods.}
\begin{figure*}[ht]
\begin{center}


  \centering
  \includegraphics[width=\textwidth]{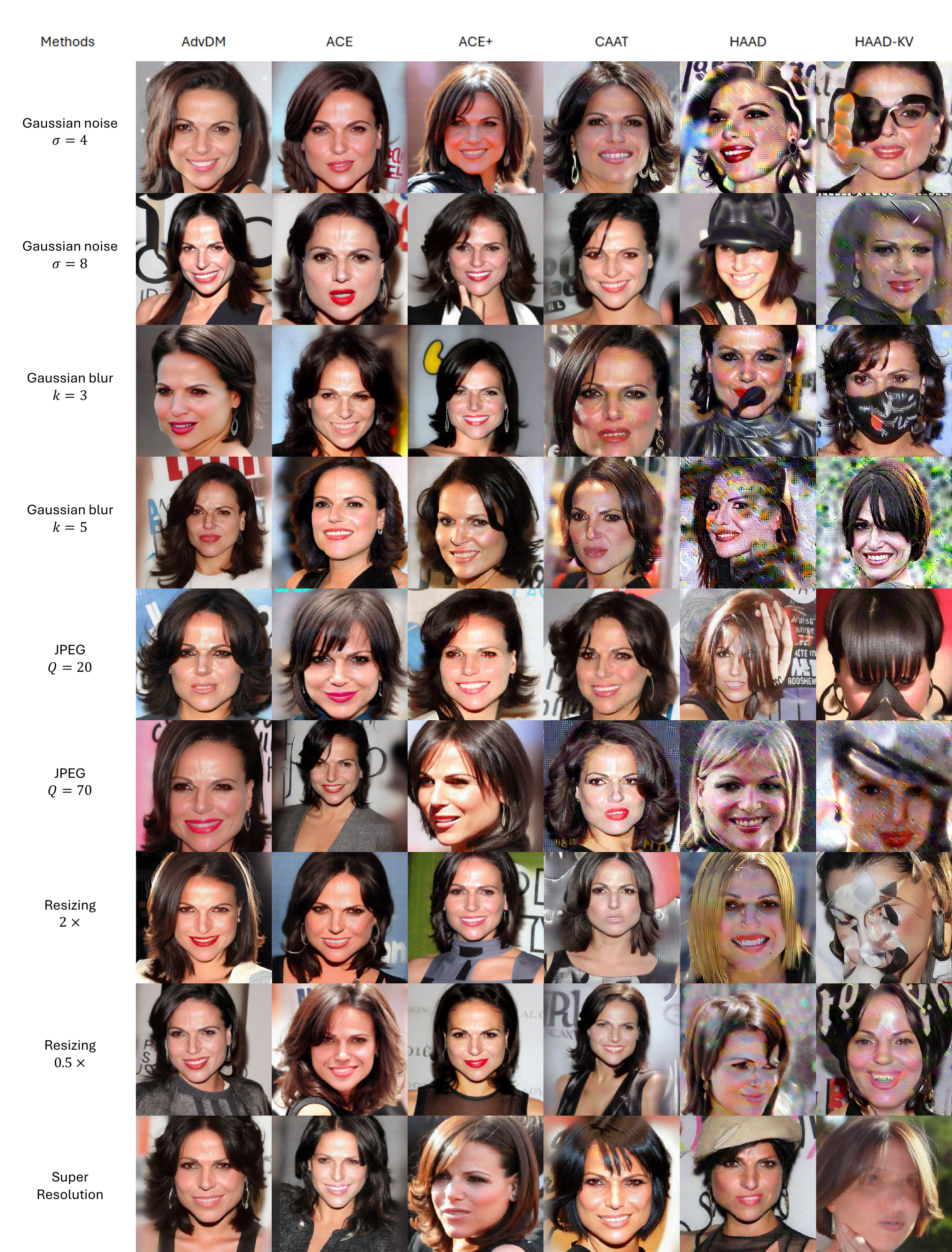}

\caption{Qualitative results for different attack methods under different purification.}
\label{app:fig:purification}
\end{center}
\vskip -0.2in
\end{figure*}

In this section, we show the effect of purification methods on the protection added by different adversarial methods. We include the various tests applied in \cite{xu2024caat,zheng2025targeted}.  
These results corroborate our conclusion: HAAD-KV not only introduces visually imperceptible perturbations but also withstands ``purification'' techniques that could otherwise neutralize fixed-pattern attacks like ACE or CAAT. Figure~\ref{app:fig:purification} provides a visual comparison of the outputs from different methods after applying the corresponding purification method. It can be seen that baseline methods often are able to recover partially recognizable content, while HAAD-KV protection consistently leads to distorted, incoherent generations — demonstrating its robustness and practical effectiveness.

\begin{table}[h]
\caption{Quantitative results for different attack methods under different purification.}
\centering
\renewcommand{\arraystretch}{0.9} 
\setlength{\tabcolsep}{5pt} 

\begin{tabular}{c|c c |c c|c c |c c|c }
\toprule
\multicolumn{10}{c}{CLIP IQA (CI) $\uparrow$} \\ \midrule
Defense    & \multicolumn{2}{c|}{Gaussian noise} 
    & \multicolumn{2}{c|}{Gaussian blur} 
    & \multicolumn{2}{c|}{JPEG} 
    & \multicolumn{2}{c|}{Resizing} 
    & \multicolumn{1}{c}{SR} \\ \midrule

Parameter & $\sigma=4$ & $\sigma=8$  & $K=3$ & $K=5$ & $Q = 20$  & $Q = 70$  & $2\times$ & $0.5\times$ &  \\ \midrule

AdvDM       & 22.71 & 20.91  & 24.88 & 29.62 & 32.91 & 21.94 & 26.23 & 22.02 & 33.23 \\ 
ACE         & 26.16 & 25.96  & 26.12 & 28.55 & 33.65 & 23.11 & 27.54 & 25.91 & 35.26 \\ 
ACE+        & 23.18 & 24.68  & 25.78 & 29.35 & 33.34 & 22.89 & 28.78 & 26.06 & 34.76 \\ 
CAAT        & \underline{28.91} & 27.81  & \textbf{31.90} & \underline{33.33} & 34.84 & 26.05 & 32.00 & \underline{26.38} & 35.55 \\
\midrule
{\bf \ourmethod} & 28.86 & \underline{28.71}  & 30.37 & 33.12 & \underline{36.18} & \underline{28.87} & \underline{33.92} & 26.37 & \underline{35.71} \\
{\bf{\ourmethod}-KV} & \textbf{31.51} & \textbf{29.15}  & \underline{31.16} & \textbf{33.61} & \textbf{39.62} & \textbf{29.26} & \textbf{34.85} & \textbf{27.88} & \textbf{36.84} \\
\bottomrule
\multicolumn{10}{c}{CLIP SIM (CS) $\downarrow$} \\ \midrule
AdvDM       & 82.41 & 81.51  & 80.24 & 78.83 & 78.94 & 80.43 & 79.18 & 81.31 & 77.72 \\ 
ACE         & 80.53 & 80.24  & 78.56 & 77.67 & 78.69 & 81.05 & 80.49 & 80.85 & 77.54 \\ 
ACE+        & 81.07 & 80.79  & 78.94 & 77.81 & 78.87 & 80.58 & 79.78 & 80.77 & 77.94 \\ 
CAAT        & 75.41 & 75.68  & \underline{75.57} & 74.67 & 78.28 & 79.96 & 78.27 & 79.39 & 76.14 \\
\midrule
{\bf \ourmethod} & \underline{74.63} & \underline{74.19}  & 75.91 & \underline{74.58} & \textbf{74.57} & \underline{78.89} & \underline{78.12} & \underline{77.03} & \underline{75.31} \\
{\bf{\ourmethod}-KV} & \textbf{68.95} & \textbf{72.53}  & \textbf{75.41} & \textbf{74.13} & \underline{75.11} & \textbf{77.91} & \textbf{76.93} & \textbf{68.98} & \textbf{72.63} \\
\bottomrule

\end{tabular}
\label{app:tab:purfication}
\end{table}

We even evaluated the robustness of our protection against recent purification techniques proposed in \cite{honig2024adversarial}, specifically Noisy Upscaling \cite{honig2024adversarial} and Impress \cite{cao2023impress}. The goal is to determine if these purification methods could remove or reduce the protection by the adversarial perturbation, thereby restoring the image's utility for personalization. Table \ref{tab:purification_results}, summarizes the impact of these methods on the image quality. We observe that, Noisy Scaling severely degrades the image quality, which is evidenced by the sharp drop in SSIM (0.3463) and PSNR (23.36). In contrast, Impress better preserves the perceptual quality, though it reduces fidelity compared to the original protected image. Notably, we used the purified images from both methods as inputs for LoRA+DreamBooth. In both cases, the personalization process failed to faithfully reconstruct the target identity. This demonstrates that HAAD-KV perturbations are robust, withstanding purification attempts and effectively preventing unauthorized customization. 

\begin{table}[h]
  \centering
  \caption{Evaluation of purification techniques against HAAD-KV protection. While purification methods like Noisy Upscaling and Impress alter image quality metrics, HAAD-KV remains effective, preventing personalization, even after purification efforts.}
  \label{tab:purification_results}
  \begin{tabular}{@{}lccc@{}}
    \toprule
    Method & SSIM $\uparrow$ & PSNR $\uparrow$ & CLIP-SIM(CS) $\downarrow$ \\
    \midrule
    No Attack (Original) & 1.0000 & $\infty$ & 83.13 \\
    HAAD-KV & 0.9862 & 59.95 & 71.91 \\
    \midrule
    \multicolumn{4}{l}{\textit{Purification applied to HAAD-KV protected Image:}} \\
    \quad Noisy Upscaling & 0.3463 & 23.36 & 77.62 \\
    \quad Impress & 0.9209 & 32.84 & 74.64 \\
    \bottomrule
  \end{tabular}
\end{table}

\clearpage

\subsection{Qualitative and quantitative results with different prompts.}

\begin{table*}[htbp]
\caption{Prompt-Invariant Performance across Six Prompts.
Quantitative results (CI, CS, FDFR, ISM) for all tested prompts, evaluating the robustness of defense methods under prompt variation. {\bf bold} is the best, while \underline{underline} indicates the second best. {\ourmethod} and {\ourmethod}-KV achieves the best performance.}
\centering
\renewcommand{\arraystretch}{0.9} 
\setlength{\tabcolsep}{5pt} 

\begin{tabular}{c|c c c c|c c c c}
\toprule
    & \multicolumn{4}{c}{"a dslr portrait of sks person"} & \multicolumn{4}{|c}{"a photo of sks person looking at the mirror"} \\ \midrule 
            & CI $\uparrow$   & CS $\downarrow$ & FDFR $\uparrow$  & ISM $\downarrow$ & CI $\uparrow$   & CS $\downarrow$ & FDFR $\uparrow$  & ISM $\downarrow$     \\
            \midrule
AdvDM       & 18.12    & 76.91     & 0.005      & 0.6366 & 23.43    & 74.75     & 0.050      & 0.5568      \\ 
ACE         & 25.96    & 74.47     & 0.010      & 0.5921 & 29.60    & 76.09     & 0.025      & 0.5923   \\ 
ACE+        & 25.77     & 75.10     & 0.007      & 0.5954 & 29.25     & 75.80     & 0.010      & 0.5908     \\ 
CAAT        & \underline{29.48}     & 74.88     & 0.060      & 0.5917 & 29.04     & 74.40     & 0.060      & 0.5388     \\
\midrule
{\bf \ourmethod}       & 29.37     & \underline{73.85}     & \underline{0.090}      & \underline{0.5899} & \textbf{31.51}     & \textbf{69.76}     & \underline{0.115}      & \textbf{0.5118}   \\
{\bf {\ourmethod}-KV}      & \textbf{30.37}    & \textbf{72.09} & \textbf{0.115}    & \textbf{0.5682} & \underline{29.76}    & \underline{72.65} & \textbf{0.145}    & \underline{0.5304}     \\
\bottomrule
    & \multicolumn{4}{c}{"a photo of sks person sitting on a chair"} & \multicolumn{4}{|c}{"a photo of sks person sitting on the floor"} \\ \midrule 
            & CI $\uparrow$   & CS $\downarrow$ & FDFR $\uparrow$  & ISM $\downarrow$ & CI $\uparrow$   & CS $\downarrow$ & FDFR $\uparrow$  & ISM $\downarrow$     \\
            \midrule
AdvDM       & 28.12    & 70.97     & 0.150      & 0.4338 & 33.24    & 66.76     & 0.105      &0.5168       \\ 
ACE         & 33.73    & 66.96     & 0.255      & 0.3944 & 34.81    & 63.67     & 0.215      &0.4728    \\ 
ACE+        & 32.57     & 67.52     & 0.220      & 0.4086 & 34.02     & 63.89     & 0.190      &0.4871      \\ 
CAAT        & 33.96     & 66.73     & 0.280      & 0.3853 & 35.15     & 62.43     & 0.260      &0.4440      \\
\midrule
{\bf \ourmethod}       & \underline{35.36}     & \underline{64.98}     &\underline{0.312}       & \underline{0.3712} & \underline{35.56}     & \underline{61.02}     & \underline{0.305}      & \underline{0.4117}   \\
{\bf {\ourmethod}-KV}      & \textbf{35.68}    & \textbf{64.13} & \textbf{0.345}    & \textbf{0.3508} & \textbf{35.82}    & \textbf{60.49} & \textbf{0.320}    & \textbf{0.4051}     \\
\bottomrule
    & \multicolumn{4}{c}{"a photo of sks person wearing glasses"} & \multicolumn{4}{|c}{"a photo of sks person talking on the phone"} \\ \midrule 
            & CI $\uparrow$   & CS $\downarrow$ & FDFR $\uparrow$  & ISM $\downarrow$ & CI $\uparrow$   & CS $\downarrow$ & FDFR $\uparrow$  & ISM $\downarrow$     \\
            \midrule
AdvDM       & 22.87    & 75.78     & 0.010      & 0.6076 & 20.78    & 78.67     & 0.025      & 0.7341      \\ 
ACE         & 23.55    & 74.21     & 0.050      & 0.5745 & 31.58    & 73.69     & 0.095      & 0.6504   \\ 
ACE+        & 23.21     & 75.02     & 0.035      & 0.5924 & 30.24     & 74.13     & 0.070      & 0.6697     \\ 
CAAT        & 23.56     & 73.75     & 0.105      & 0.5554 & 32.81     & 73.43     & 0.115      & 0.6332     \\
\midrule
{\bf \ourmethod}       & \underline{23.99}     & \underline{73.06}     & \underline{0.155}      & \underline{0.5302} & \underline{33.43}     & \underline{72.56}     & \underline{0.190}      & \underline{0.6279}   \\
{\bf {\ourmethod}-KV}      & \textbf{24.18}    & \textbf{72.87} & \textbf{0.185}    & \textbf{0.5217} & \textbf{33.51}    & \textbf{71.78} & \textbf{0.235}    & \textbf{0.6261}     \\
\bottomrule

\end{tabular}
\label{app:tab:prompt}
\end{table*}

\clearpage

\subsection{Transferability study using HAAD : Qualitative results.}

\begin{table}[htbp]
\caption{Transferability of {\ourmethod} across different versions of SD.}
    \label{tab:transferability-sd-version-qual}
    \centering
    \renewcommand{\arraystretch}{0.9} 
    \setlength{\tabcolsep}{5pt} 
    \begin{tabular}{c c c c c c c c c c}
            \hline
            Target      & \multicolumn{3}{c}{SD1.4} & \multicolumn{3}{c}{SD1.5}  & \multicolumn{3}{c}{SD2.1}  \\ \hline
            Attacker    & LoRA+DB & \multicolumn{2}{c}{SDEdit} & LoRA+DB & \multicolumn{2}{c}{SDEdit} & LoRA+DB & \multicolumn{2}{c}{SDEdit} \\ 
                        & CI $\uparrow$    & MS $\downarrow$     & CS $\downarrow$  & CI $\uparrow$   & MS $\downarrow$        & CS $\downarrow$        & CI $\uparrow$      & MS $\downarrow$       & CS $\downarrow$ \\ \midrule
            
            No Attack   & 18.89 & 0.3694 & 75.01  & 21.32   & 0.3637    & 79.26     & 19.18     & 0.3782    & 74.67
\\ 
            
            SD1.4       & 27.51 & 0.3498 & 73.56  & 29.38   & 0.3444    & 73.62     & 30.36     & 0.3587    & 72.93
\\ 
            
            SD1.5       & 29.32 & 0.3479 & 73.23  & 29.10   & 0.3360    & 75.80     & 30.55     & 0.3568    & 72.85
\\ 
            
            SD2.1       & 29.27 & 0.3455 & 73.18  & 26.81   & 0.3435    & 73.23     & 29.62     & 0.3561    & 72.72
\\ \hline

        \end{tabular}  
\end{table}

\label{app:transferability}

\begin{table*}[!h]
\caption{Transferability of {\ourmethod} among the different versions of SD. We observe consistent degradation for all the methods.}
    \label{app:tab:transferability-sd-version-visual}
    \centering
    \renewcommand{\arraystretch}{1.5}
        \begin{tabular}{ >{\centering\arraybackslash}m{0.1\textwidth} 
                 >{\centering\arraybackslash}m{0.11\textwidth} 
                 >{\centering\arraybackslash}m{0.12\textwidth} 
                 >{\centering\arraybackslash}m{0.11\textwidth} 
                 >{\centering\arraybackslash}m{0.12\textwidth} 
                 >{\centering\arraybackslash}m{0.11\textwidth} 
                 >{\centering\arraybackslash}m{0.12\textwidth} }
            \toprule
            Target      & \multicolumn{2}{c}{SD1.4} & \multicolumn{2}{c}{SD1.5}  & \multicolumn{2}{c}{SD2.1}  \\ 
            \midrule
            Attacker    & LoRA+DB & SDEdit & LoRA+DB & SDEdit & LoRA+DB & SDEdit \\ 
            \toprule
            
            No Attack & 
            \includegraphics[width=0.13\textwidth]{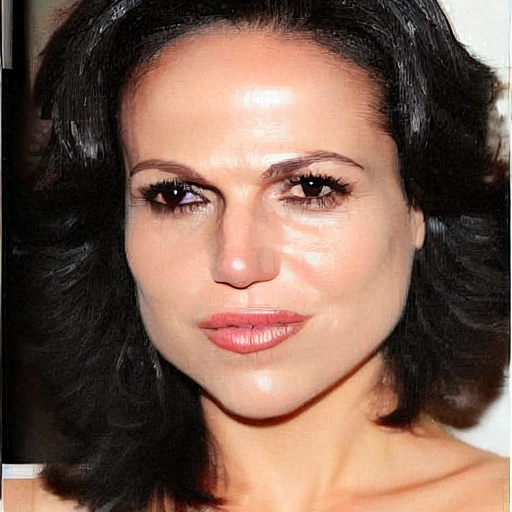} & 
            \includegraphics[width=0.13\textwidth]{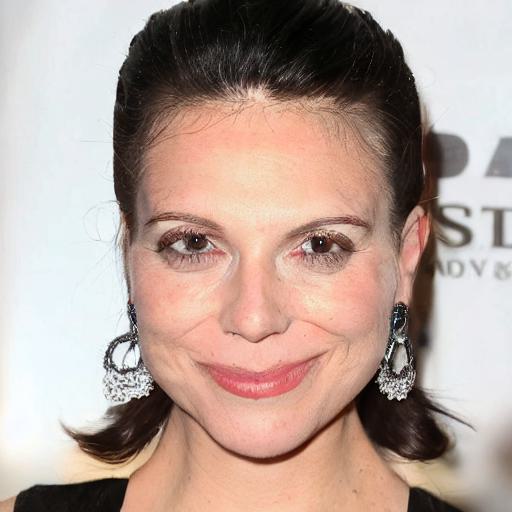} & 
            \includegraphics[width=0.13\textwidth]{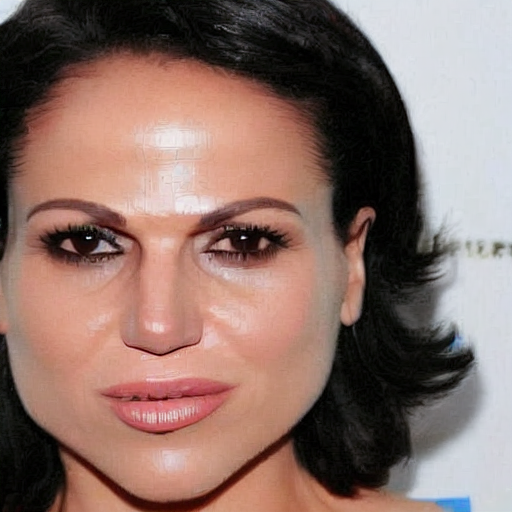} & 
            \includegraphics[width=0.13\textwidth]{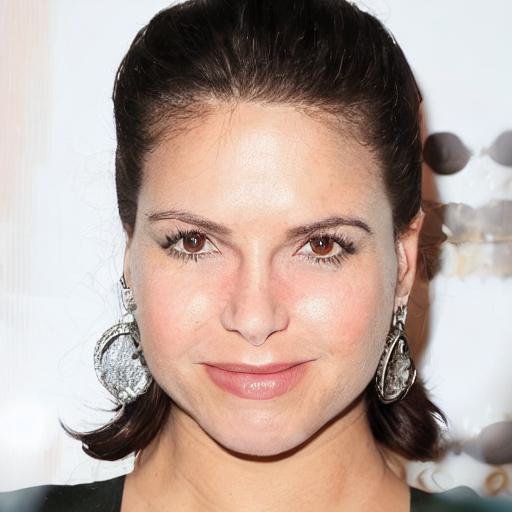} & 
            \includegraphics[width=0.13\textwidth]{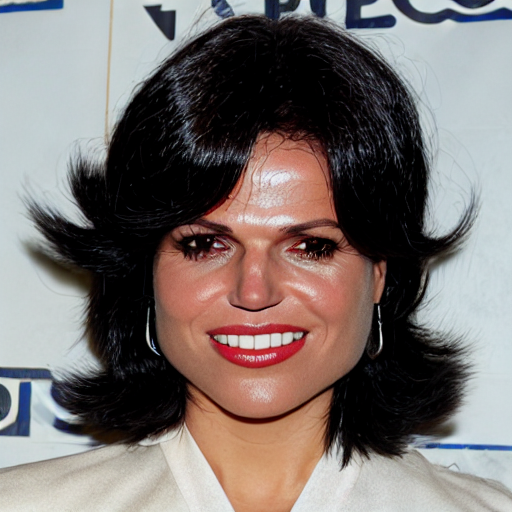} & 
            \includegraphics[width=0.13\textwidth]{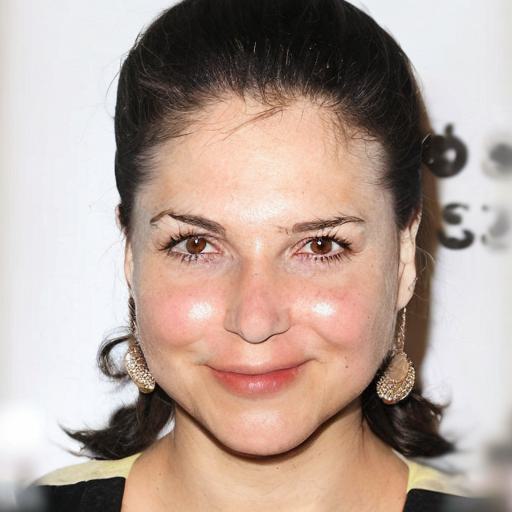} \\ 
            
            SD1.4 & 
            \includegraphics[width=0.13\textwidth]{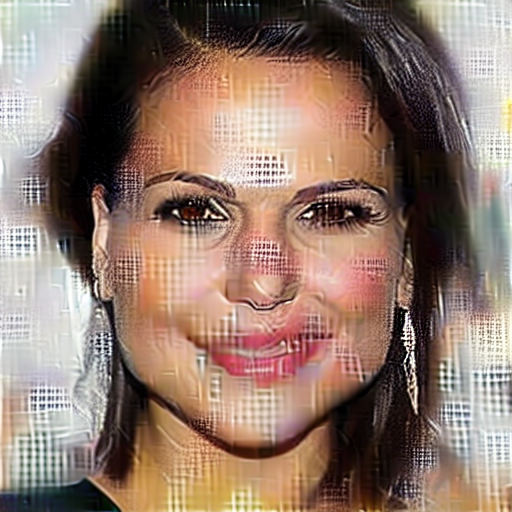} & 
            \includegraphics[width=0.13\textwidth]{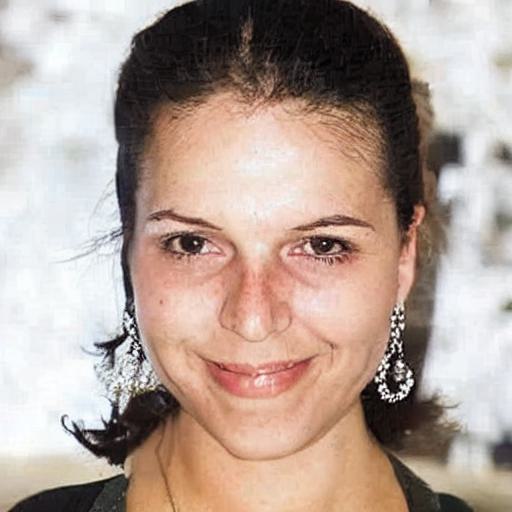} & 
            \includegraphics[width=0.13\textwidth]{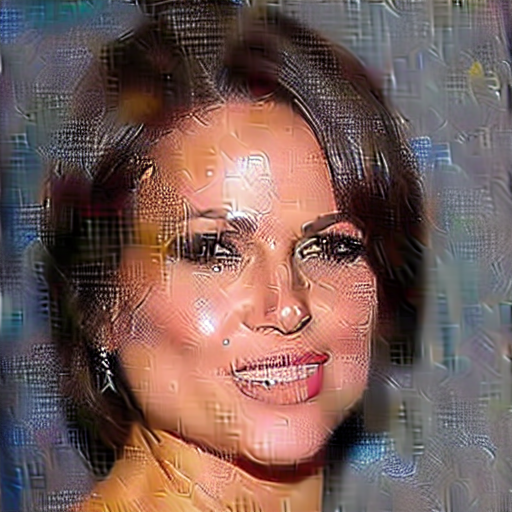} & 
            \includegraphics[width=0.13\textwidth]{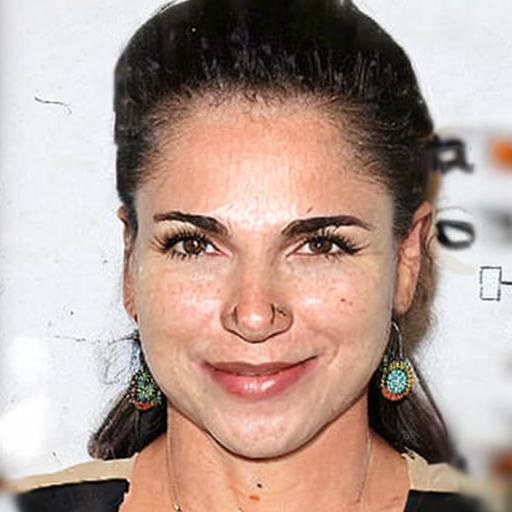} & 
            \includegraphics[width=0.13\textwidth]{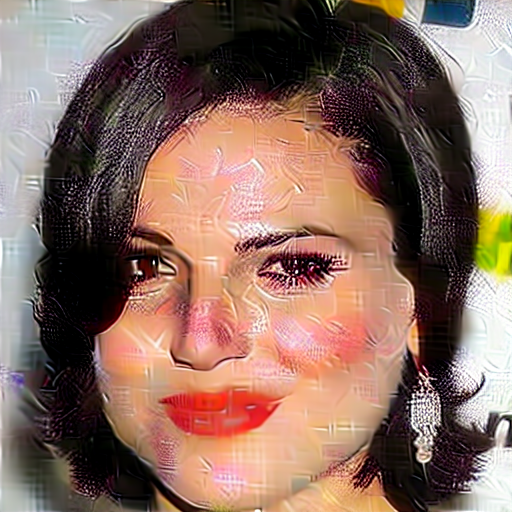} & 
            \includegraphics[width=0.13\textwidth]{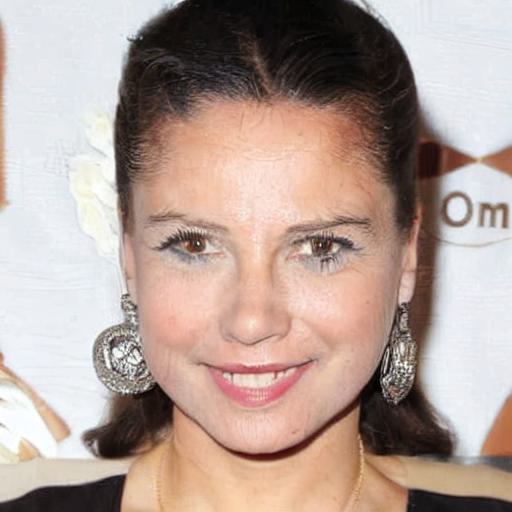} \\ 
            
            SD1.5 & 
            \includegraphics[width=0.13\textwidth]{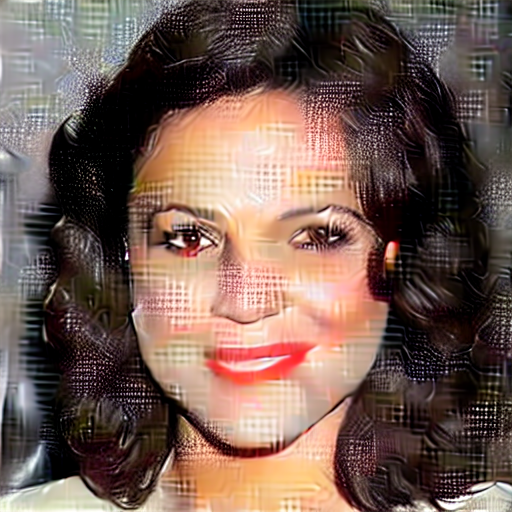} & 
            \includegraphics[width=0.13\textwidth]{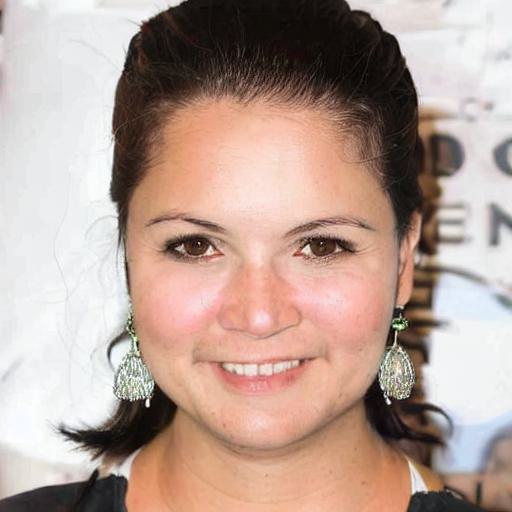} & 
            \includegraphics[width=0.13\textwidth]{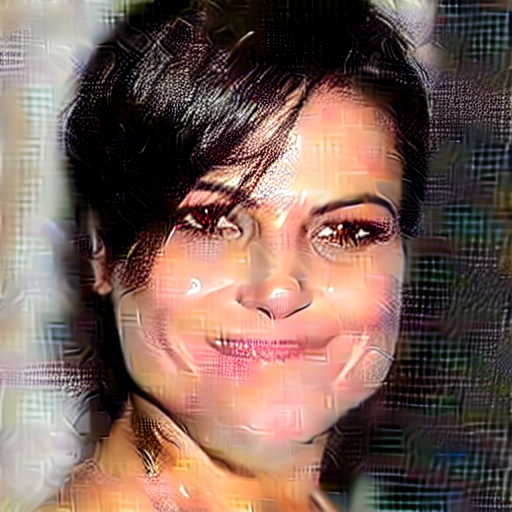} & 
            \includegraphics[width=0.13\textwidth]{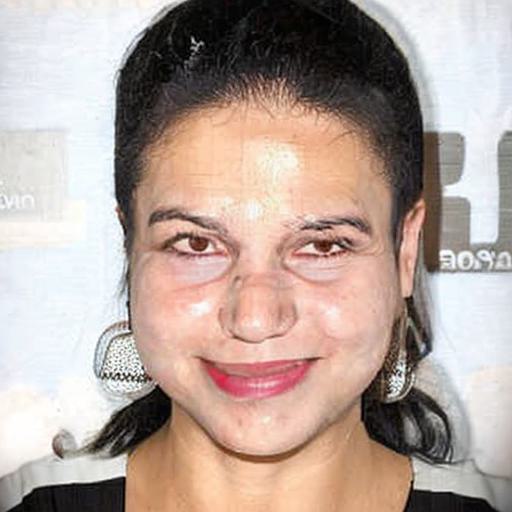} & 
            \includegraphics[width=0.13\textwidth]{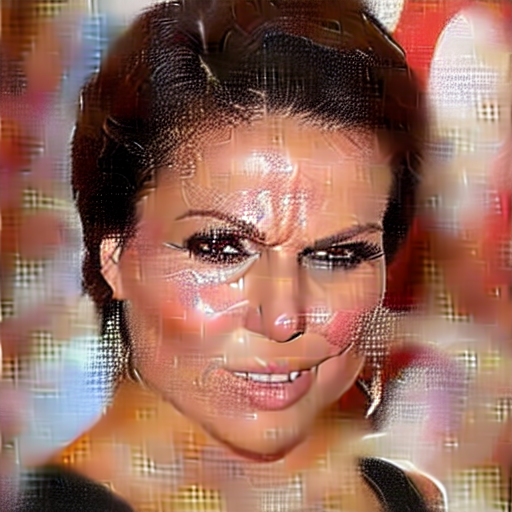} & 
            \includegraphics[width=0.13\textwidth]{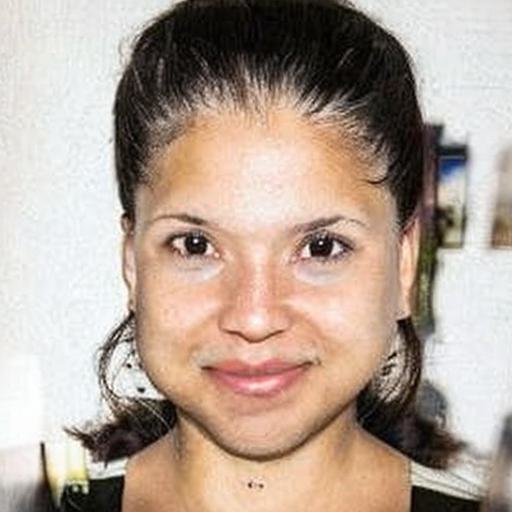} \\ 
            
            SD2.1 &
            \includegraphics[width=0.13\textwidth]{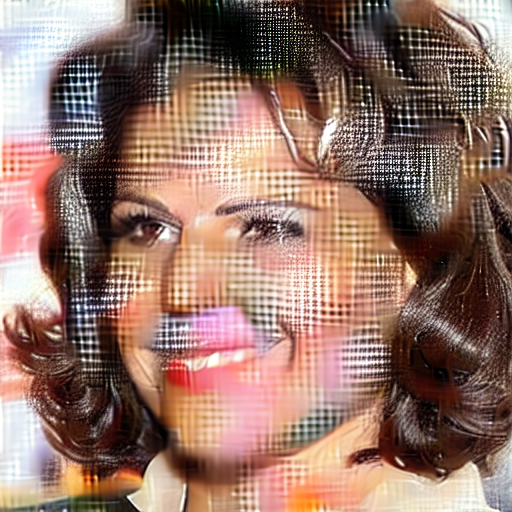} & 
            \includegraphics[width=0.13\textwidth]{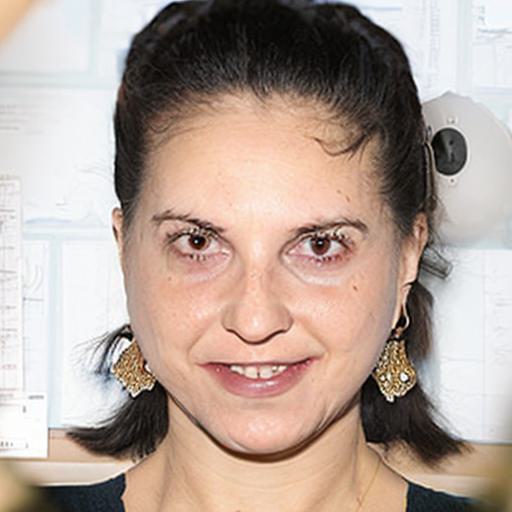} & 
            \includegraphics[width=0.13\textwidth]{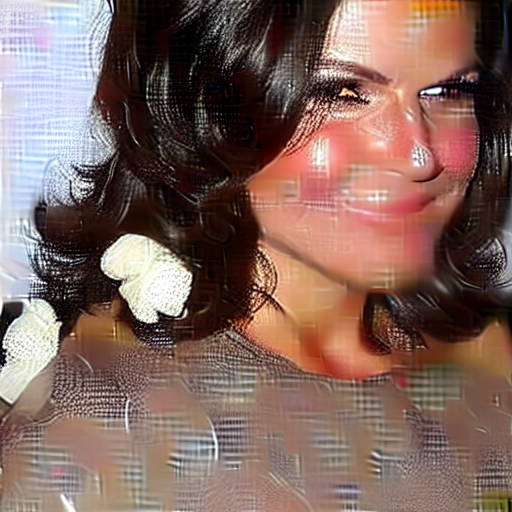} & 
            \includegraphics[width=0.13\textwidth]{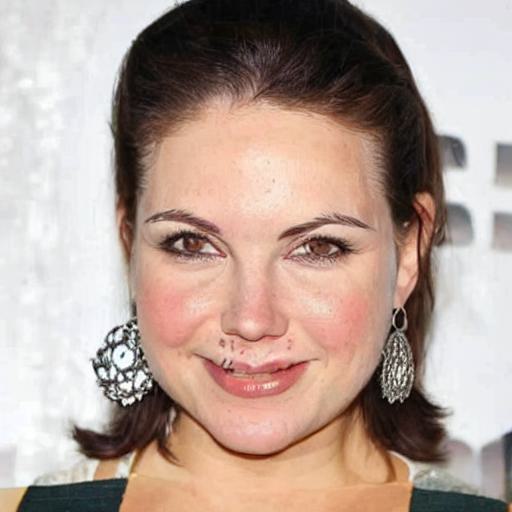} & 
            \includegraphics[width=0.13\textwidth]{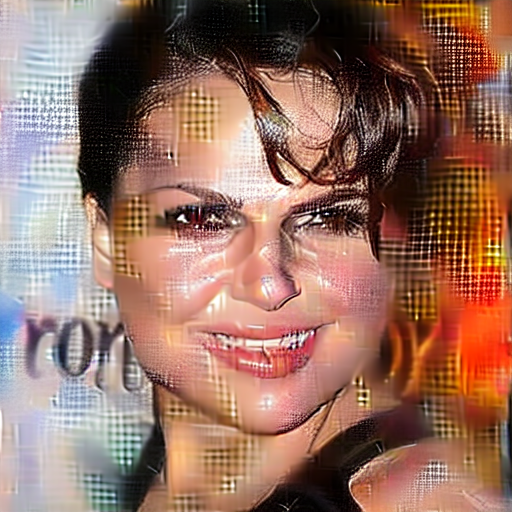} & 
            \includegraphics[width=0.13\textwidth]{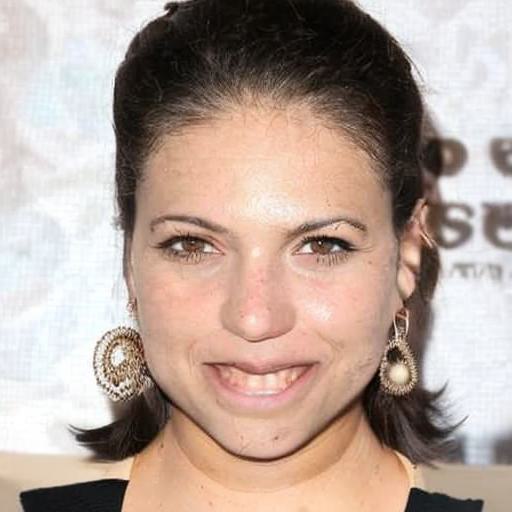} \\ 
            \bottomrule
            
        \end{tabular}    
\end{table*}

\clearpage
\subsection{Visualization of cross-attention maps for all the attack methods.} 
\label{app:haad-kv-block-visual}

\clearpage
\subsection{Theoretical Framework.}
\label{app:sec:theoretical}
\textbf{\textit{h}-space Perturbation and Semantic Misalignment.} Let $W_h$ denote the features (weights) of the \textit{h}-space of the U-Net, which encodes high-level semantic content during the diffusion process. For the HAAD attack, we iteratively compute adversarial perturbation for the protection of the input image $x$ with projected gradient descent(PGD) which maximizes the reconstruction loss $\mathcal{L}_{\text{LDM}}$. Each step updates the semantic features as given by:

\begin{equation}
W_h' = W_h + l \nabla_{W_h} \mathcal{L}_{\text{LDM}}(x),
\end{equation}

where, $l$ is the learning rate. The final protected image is denoted as $x' = x + \delta$, and its corresponding semantic features are encoded in $W_h'$.

This update induces a linear shift in the \textit{h}-space, modifying the intermediate representation in a direction aligned with $\nabla_{W_h}$, which was adversarially chosen to degrade the semantic consistency.  Since the cross-attention mechanism depends on the alignment between the textual prompt and the latent features in the \textit{h}-space, this shift weakens the model’s ability to associate prompt tokens (e.g., “sks”) with relevant visual features, which finally affects the image generation process.

When an attacker trains a new personalization model with the protected image
\( x' \), it starts to operate on the clean semantic representation encoded in \( W_h\). Afterwards, at the end of the personalization process, the resulting \textit{h}-space parameters are denoted as \( W_h'' \). Because the training began with $x'$ and \( W_h\), we expect: $W''_h \approx W'_{h}$, implying that the learned representation converges towards a similar semantically misaligned features (weights) ($W'_{h}$). This is straightforward since $x'$ carries the semantic misalignment information encoded in the perturbation.

Consequently, at inference time, the model’s attention fails to bind the user-specific token (e.g., “sks”) to the correct semantic features, due to the distortion introduced by the perturbation \( \Delta W_h = W_h'' - W_h \). This semantic misalignment in the attention mechanism results in visibly degraded or incoherent image generations. Thus, perturbations in \textit{h}-space not only alter intermediate representations but also disrupt the prompt-to-concept mapping, offering an effective and semantically grounded defense against unauthorized personalization.

\textbf{The role of KV layers towards Perturbation and Semantic Misalignment.} HAAD-KV extends the above intuition by focusing the adversarial update specifically on the \textit{cross-attention layers} inside the \textit{h}-space block, and only modifying the key $W_K$ and value $W_V$ projection matrices.

During few-shot personalization, these matrices adapt to encode how each visual element should respond to the tokenized concept. HAAD-KV injects perturbations only into $W_K$ and $W_V$, thereby corrupting the model’s ability to compute meaningful attention maps. In other words, the perturbed matrices $W'_K$ and $W'_V$ are given by:
\[
W'_K=W_K+l \nabla_{W_K} \mathcal{L}_{\text{LDM}}(x),
\]
\[
W'_V=W_V+l \nabla_{W_V} \mathcal{L}_{\text{LDM}}(x)
\]
This formulation is similar to equation (3). As a result, the perturbed attention becomes:

\[
\text{Attention}(Q, K', V') = \text{softmax}\left( \frac{QK'^\top}{\sqrt{d}} \right)V'
\]

where, the $K'$ and $V'$ are the resultant perturbed matrices, induced by  $W'_K$ and $W'_V$, respectively. Since this misalignment occurs at the point where identity tokens are explicitly bound to visual representations, the model fails to personalize effectively during training. 

In conclusion, HAAD-KV thus offers a focused and computationally efficient attack: by perturbing only a small fraction ($\sim$5\% of the total parameters in \textit{h}-space) of model parameters (those with high semantic content), it results in maximum distortion of the personalization mechanism with minimal perceptual footprint.

\textbf{Analysis of representational disruption in \textit{h}-space.} To quantitatively evaluate that our method alters the core semantic structure, we perform a Principal Component Analysis (PCA) on the activations within the \textit{h}-space. This analysis aims to measure the alignment of the primary directions of variance between representations of clean and protected images. We computed the cosine similarity between the top-k principal components derived from clean image activations and those from images protected by HAAD-KV.

\noindent \textit{Description of the PCA method: }PCA identifies the orthogonal directions, or Principal Components (PCs), that capture the maximum variance within the \textit{h}-space activations. The ``top-k components'' refer to the first $k$ PCs that represent the most significant semantic patterns in the data. We then use cosine similarity to measure the alignment between the PCs derived from clean images  with those from their protected counterparts. A cosine similarity value close to 1.0 implies that the components are highly aligned, meaning the clean and protected images share the same core semantic structure. Conversely, a value close to 0.0 indicates orthogonality, meaning their semantic structures are fundamentally different and uncorrelated.

The results presented in Table \ref{tab:pca_analysis}, clearly reveal a systematic and noise budget dependent impact of HAAD-KV. At a low perturbation budget of $\eta=4/255$, we already observe a significant drop in alignment, with the cosine similarity for the top-5 components being 0.2961. This indicates that even at low perturbation budget the protection begins to alter the primary semantic features. As the perturbation budget increases to $\eta=8/255$ and $\eta=16/255$, respectively, this trend is further amplified. For instance, at $\eta=16/255$, the similarity of the top-5 components drops to 0.1382. This demonstrates that larger protection budgets lead to a greater divergence in semantic representation. Furthermore, this effect is consistent when more components were considered (i.e., k is increased from 5 to 50), confirming that the semantic shift is not limited to only the top few dominant features but across the semantic latent space.

\begin{table}[h]
  \centering
  \caption{Cosine similarity between the top-$k$ principal components of h-space activations for clean versus HAAD-KV protected images. The consistently low similarity scores across different perturbation budgets ($\eta$) demonstrate a significant semantic drift induced by HAAD-KV.}
  \label{tab:pca_analysis}
  \begin{tabular}{@{}ccccc@{}}
    \toprule
    Perturbation ($\eta$) & \multicolumn{4}{c}{Top-$k$ Principal Components} \\
    \cmidrule(l){2-5}
     & $k=5$ & $k=10$ & $k=20$ & $k=50$ \\
    \midrule
    $4/255$  & 0.2961 & 0.1413 & 0.0668 & 0.0243 \\
    $8/255$  & 0.2171 & 0.1111 & 0.0603 & 0.0245 \\
    $16/255$ & 0.1382 & 0.0672 & 0.0351 & 0.0135 \\
    \bottomrule
  \end{tabular}
\end{table}

\subsection{User study.}
\label{app:sec:user_study}

To empirically verify that the injected noise is imperceptible to humans, we carried out a user study with the CelebA-HQ dataset, containing face images. For each noise budget $\eta \in \{4/255, 8/255, 16/255\}$ we sampled \textbf{10} identities, and for every identity selected \textbf{3} high-resolution photographs, yielding \textbf{30} original--perturbed pairs per $\eta$ value. Twenty-six volunteers (\textbf{26}$\times$30 = 780 judgements per budget) were shown each pair side-by-side and asked \emph{``Which image is perturbed?''} with three choices: \textit{(A)} first, \textit{(B)} second, or \textit{(C)} ``both images are the same.''

Table \ref{tab:human} reports the scores obtained for the three noise budgets, including the \textbf{Error Rate} and \textbf{Z-score} derived from Thurstone's Case V model\cite{thurstone1927scale}. In this study, the Error Rate represents the fraction of times volunteers failed to correctly identify the perturbed image (Accuracy(\%) is given by \{100-(Error Rate*100)\}\%). A higher error rate signifies that the introduced noise is less perceptible(imperceptible) and is hard to be detected by human eyes. The Z-score quantifies perceptual discriminability, with a higher value indicating easier detection. The \textbf{Standard Deviation (STD)} of the Z-score measures the level of agreement among the participants, where a lower STD indicates a more consistent perceptual experience across the group.

The analysis reveals a strong correspondence between the metrics. Specifically, for $\eta=16/255$, a high positive Z-score of ``1.23'' indicates that the perturbation was clearly perceptible, a conclusion strongly supported by the very low error rate of ``0.117'' (88.3\% accuracy) and low STD (0.29). And the low STD (0.29) for this budget shows a strong consensus that many were able to detect the perturbation in the image. For $\eta=8/255$, the Z-score is near zero (``0.28'') suggesting discrimination was at a threshold level with a mix of both partial detection and low perception, which is corroborated by the high STD (``0.34'') and a moderate error rate of ``0.392'' (60.8\% accuracy). Critically, for $\eta=4/255$, the negative Z-score of ``-0.39'' confirms the images were perceptually indistinguishable. This is strongly demonstrated by the error rate of ``0.650'', which translates to an accuracy of 35\%—statistically identical to random chance for this 3-choice task. The low STD (``0.25'') for this budget further shows a strong consensus that no difference could be spotted.
This confirms that at the strictest budget ($4/255$), participants performed at chance level, indicating that the perturbations are \emph{visually indistinguishable} in practice.
\vspace{-0.33cm}
\begin{table}[h]
    \centering
    \begin{tabular}{lccc}
        \toprule
        \textbf{Noise budget $\eta$} & \textbf{Mean Z-Score} & \textbf{STD} & \textbf{Error Rate} \\
        \midrule
        $4/255$  & -0.39 & 0.25 & 0.650 \\
        $8/255$  & 0.28  & 0.34 & 0.392 \\
        $16/255$ & 1.23  & 0.29 & 0.117 \\
        \bottomrule
    \end{tabular}
    \caption{Thurstone Case V results and corresponding error rates. The Z-scores and error rates provide complementary evidence for the (in)discriminability of perturbations.}
    \label{tab:human}
\end{table}

\clearpage
\subsection{Qualitative results for different attack methods under different customization models.} 
\label{app:compare-methods}

\end{document}